\documentclass[11pt]{article}

\usepackage[final]{acl}

\usepackage{times}
\usepackage{latexsym}

\usepackage[T1]{fontenc}

\usepackage[utf8]{inputenc}

\usepackage{microtype}

\usepackage{inconsolata}

\usepackage{graphicx}

\usepackage{pgfplots}
\pgfplotsset{compat=1.17}
\usepgfplotslibrary{groupplots}
\usetikzlibrary{positioning,calc}

\usepackage{booktabs}

\usepackage{amsmath}
\usepackage{amssymb}
\usepackage{multirow}
\newcommand{\affilsup}[1]{#1}

\title{FineWeb-CLaR: Culture, Language, and Region Annotations for Benchmark-Aligned Corpus Auditing}

\author{
  Yusser Al Ghussin\affilsup{$^{1,2}$}  \qquad Eva Gavaller\affilsup{$^{1,2}$}  \\ 
  \textbf{Cristina España-Bonet}\affilsup{$^{2,3}$} \qquad \textbf{Josef van Genabith}\affilsup{$^{1, 2}$} \qquad \textbf{Simon Ostermann}\affilsup{$^{1,2}$} \\
  \\
  \small $^1$ Saarland University \enspace $^2$ German Research Center for Artificial Intelligence (DFKI) \\ 
   \enspace \small $^3$ Barcelona Supercomputing Center (BSC-CNS), Barcelona, Catalonia, Spain
   \\
  \texttt{\small yusser.al\_ghussin@dfki.de}
}

\begin{document}
\maketitle

\begin{abstract}
Cultural evaluation coverage and robustness in language models are difficult to diagnose because pretraining corpora and cultural benchmarks are rarely indexed with comparable metadata. Benchmarks increasingly target culturally situated phenomena at the level of languages, regions, and locale-specific practices, while web-scale corpora are usually organized only by language. A shared culture-language-region layer makes these resources comparable, enabling audits of whether a target cultural phenomenon is represented in pretraining data, evaluated by benchmarks or both. To this end, we introduce FineWeb-CLaR, a large-scale annotated dataset derived from FineWeb and FineWeb-2 that places web documents on a shared culture--language--region axis for corpus auditing and benchmark alignment.

FineWeb-CLaR annotates the full 30.9B-document collection from FineWeb and FineWeb-2 with URL-derived region labels and cultural-topic provenance. Our region resolver assigns a non-empty region to 25.61\% of documents (7.92B). For cultural-topic analysis, we induce locale-specific topics and project them onto the 14 leaves of the Cultural Taxonomy of \citet{liu2025culturally}, producing Locale Topic Distributions (LTDs) for corpus-side comparison. We also annotate 277 cultural NLP benchmarks with the same taxonomy, language coverage, and region coverage. Together, these resources enable direct comparison between corpus-side pretraining evidence and benchmark-side evaluation coverage.\footnote{Code and data: \\ \url{https://github.com/Yusser96/FineWeb-CLaR} \\ \url{https://huggingface.co/collections/Yusser/culture-datasets}}
\end{abstract}

\section{Introduction}
\label{sec:introduction}

Cultural evaluation coverage and robustness in language models are difficult to diagnose because failures may stem from missing cultural evidence in pretraining data, weak benchmark coverage, poor alignment with target communities, or model behaviour that relies on surface associations rather than culturally grounded understanding~\citep{adilazuarda2024towards,chiu2024culturalbench,naous-xu-2025-origin,vo2025cure}. Diagnosing these cases requires comparing the cultural phenomena targeted by benchmarks with the evidence available in pretraining data.

Existing web-scale pretraining corpora are not designed for this comparison. Corpora such as FineWeb \citep{penedo2024fineweb} and FineWeb-2 \citep{penedo2025fineweb2} are organized primarily by detected language, but language is too coarse for culturally situated analysis: Many languages span multiple regions, institutions, communities, and communicative norms. Following the Cultural Taxonomy (CT) of \citet{liu2025culturally}, we treat culture as \textit{socially shared phenomena} organized into ideational, linguistic, and social dimensions, covering knowledge, values, beliefs, norms, language varieties, social groups, and relations. Existing language-only corpus annotations cannot show which of these phenomena appear in which regional or locale-specific slices of pretraining data.

To that end, we introduce FineWeb-CLaR, a large-scale annotated dataset derived from FineWeb and FineWeb-2 that places web documents on a shared language-region-culture
axis. The language axis comes from the original corpus; for the region axis we assign URL-derived ISO~3166-1 alpha-2 labels with confidence metadata; and for the cultural-labels axis we map the corpus to the 14 leaves of the CT via a topic-based rule set.
Our contributions are:


\begin{itemize}
    \item We introduce FineWeb-CLaR, a large-scale version of FineWeb and FineWeb-2 augmented with region metadata and CT-aligned topic metadata for culture-, language-, and region-aware corpus auditing.

    \item We propose a scalable URL-based region attribution method for web-scale data with confidence and resolution-source metadata.

    \item We derive cultural-taxonomy-aligned locale-level data using topic modeling, embedding-based ranking and LLM adjudication.

    \item We extend prior taxonomy-driven cultural benchmark surveys~\citep{liu2025culturally,namazifard2025cultureneurons} by annotating 212 additional cultural NLP benchmarks with locale metadata.
\end{itemize}

\section{Related Work}
\label{sec:related-work}

\paragraph{Culture, benchmarks, and corpus audits.}
Recent work has argued that NLP systems and evaluations often conflate language, nationality, ethnicity, geography, and culture, while cultural evaluation resources vary in how they operationalise knowledge, values, norms, social relations, and situated behaviour \citep{hershcovich2022challenges,adilazuarda2024towards,liu2025culturally}. We build on the Cultural Taxonomy of \citet{liu2025culturally}, which organizes cultural phenomena into ideational, linguistic, and social dimensions, and use it as a shared coding scheme for corpus-side topic metadata and benchmark-side target phenomena. 
Prior benchmark efforts evaluate multicultural factual knowledge, everyday cultural knowledge, values, social bias, stereotypes, and culturally grounded reasoning \citep{myung2024blend,chiu2024culturalbench,durmus2023globalopinions,pistilli2024civics,nangia2020crows,parrish2022bbq}. Taxonomy-driven resource surveys and curations, including \citet{liu2025culturally} and MUREL \citep{namazifard2025cultureneurons}, make these resources more comparable. Our benchmark audit follows this line of work, but connects benchmark coverage to corpus-side evidence on a pretraining scale, by annotating both with the same language--region-taxonomy metadata.

\paragraph{Web corpora and geographic attribution.}
Large web corpora such as CCNet, OSCAR, mC4, CC100, CulturaX, FineWeb, and FineWeb-2 improve pretraining data transparency through language identification, filtering, deduplication, and quality-control pipelines, but they are primarily organized by detected language rather than by region or locale \citep{wenzek2020ccnet,ortizsuarez2020monolingual,xue2021mt5,conneau2020unsupervised,nguyen2023culturax,penedo2024fineweb,penedo2025fineweb2}. Geographically organized web corpora are a step closer to our culture–language–region annotation objectives: the Corpus of Global Language Use indexes Common Crawl data by language-country pairs \citep{dunn2020mapping}, and GloWbE organizes web English into country-specific varieties \citep{davies2015glowbe}. We follow this tradition of URL-based geographic attribution, using ccTLDs, public-suffix-aware domain parsing, host and path tokens, and locale parameters. Because such signals are noisy and do not directly identify the author's location, audience, or cultural membership, we release region labels together with confidence and resolution-source metadata.

\paragraph{Topic modeling and cultural mapping.}
Topic modeling is widely used to summarize large corpora into interpretable thematic distributions, from classical models such as LDA \citep{blei2003lda} to neural methods such as BERTopic and FASTopic \citep{grootendorst2022bertopic,wu2024fastopic}. Multilingual topic models have also been used to compare culturally or linguistically distinct corpora \citep{gutierrez-etal-2016-detecting}. Our use of topic modeling differs from work that treats discovered topics as the final analysis object, and from supervised label-aligned topic models that use task labels during training \citep{yang2025labelalignment}. Instead, we induce topics independently within each language-region locale using multilingual document BGE-M3 embeddings \citep{chen2024m3}, and FASTopic \citep{wu2024fastopic}, then project topic centroids post hoc onto the 14 leaves of the Cultural Taxonomy to obtain locale topic distributions.

\section{FineWeb-CLaR Corpus Annotation}
We describe our two-stage annotation approach for labeling FineWeb \citep{penedo2024fineweb} and FineWeb-2 \citep{penedo2025fineweb2} with region information and mapping the resulting language-region locales to the Cultural Taxonomy of \citet{liu2025culturally}.

\subsection{Region Attribution}
\label{sec:region-attribution}

Cultural benchmarks often specify evaluation targets at the level of countries, regions, dialect areas, or language--region communities. Such annotations are usually missing in pretraining data, which are organized primarily by language. We close this gap and add document-level region metadata to FineWeb and FineWeb-2 so that pretraining evidence can be aggregated into units closer to those used by cultural evaluations.

\subsubsection{Region Labels}
\label{sec:region-definition}

A region label can refer to several different concepts, including author location, server location, target audience, domain jurisdiction, or the country conventionally associated with a language. We use a narrower operational definition: for a URL $u$, a region label $r$ means that URL-derived evidence supports an association between $u$ and the ISO~3166-1 alpha-2 region $r$ \citep{iso3166}. Prior work has shown that web-scale corpora can be organized geographically using URL-, domain-, and TLD-derived evidence \citep{dunn2020mapping}, but ccTLD and public-web signals provide only partial and source-dependent coverage of local web spaces \citep{sommese2023local}. We therefore use URL-derived region labels only as weak auditing metadata, not as author-location, audience-location, or cultural-membership labels.

This narrow definition is necessary because URL evidence is scalable but noisy. URLs and crawl metadata are available for web-scale corpora and are preserved in Common Crawl WARC/WAT records \citep{commoncrawl2023formats}. Regional signals such as ccTLDs, hostnames, path tokens, query parameters, and language headers can provide evidence of regional association, but they are imperfect: domains may be used outside their nominal jurisdictions, generic TLDs such as \texttt{.com} are widespread, and hosting or domain choices do not directly identify authorship or cultural affiliation. We therefore release region labels together with source and confidence metadata.

\begin{table}[h]
\centering
\small
\begin{tabular}{llll}
\hline
\textbf{\#} & \textbf{Source} & \textbf{Signal} & \textbf{Conf} \\
\hline
1 & \texttt{query} & explicit locale parameter & \texttt{high} \\
2 & \texttt{host\_path} & lookup table & inherited \\
3 & \texttt{host} & lookup table & inherited \\
4 & \texttt{domain} & lookup table & inherited \\
5 & \texttt{url\_hint} & curated URL token & \texttt{url\_hint} \\
6 & \texttt{tld} & non-branded ccTLD & \texttt{ccTLD} \\
7 & \texttt{none} & no region signal & \texttt{none} \\
\hline
\end{tabular}
\caption{Search-time priority chain for region resolution, with lookup-backed sources inheriting table confidence (high, medium or low) and fallback sources receiving weak confidence labels.}
\label{tab:region-resolution-priority}
\end{table}

\subsubsection{Resolver Design}
\label{sec:pipeline-overview}

Our region resolver separates offline region evidence aggregation from corpus-scale annotation and emits three fields for each document: a region label, a resolution source, and a confidence value. The region label is an ISO~3166-1 alpha-2 code or \texttt{XX} when no region can be resolved; the resolution source records which URL signal produced the label; and the confidence value records the resolver's confidence in that assignment.

In the offline stage, we build a frozen URL-signature lookup table from Common Crawl records sampled across the available \texttt{CC-MAIN} archives, ranging from \texttt{CC-MAIN-2008-2009} to \texttt{CC-MAIN-2026-08}. Each sampled record contributes a URL and, when available, HTTP \texttt{Content-Language}, HTML \texttt{lang}, and language-identification metadata from FineWeb. We extract URL signatures at three granularities: domain, host, and host plus first path segment, then aggregate weighted evidence over candidate regions for each signature. The following example illustrates the terminology:
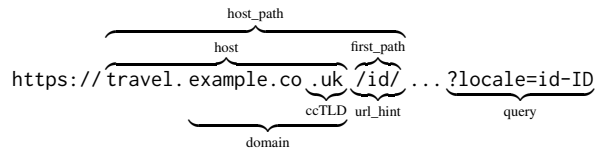
\[
\resizebox{\linewidth}{!}{$
\texttt{https://}
\overbrace{\overbrace{
  \texttt{travel.}
  \underbrace{
    \texttt{example.co}
    \underbrace{\texttt{.uk}}_{\text{\scriptsize ccTLD}}
  }_{\text{\scriptsize domain}}
}^{\text{\scriptsize host}}
\underbrace{
  \overbrace{\texttt{/id/}}^{\text{\scriptsize first\_path}}
}_{\text{\scriptsize url\_hint}}}^{\text{\scriptsize host\_path}}
\texttt{...}
\underbrace{\texttt{?locale=id-ID}}_{\text{\scriptsize query}}
$}
\]

Each lookup row stores the winning region, confidence tier, and supporting metadata; details of the evidence rules are provided in Appendix~\ref{app:region-resolver-details}.

In the search-time stage, we apply the frozen lookup table to every FineWeb and FineWeb-2 document. Given a URL, the resolver follows the fixed priority chain shown in Table~\ref{tab:region-resolution-priority}; the first source that emits a valid label determines the document's region and resolution source. Lookup-backed matches inherit the confidence metadata stored in the frozen lookup table, while fallback sources receive separate weak confidence labels. Details of the search-time resolution sources are provided in Appendix~\ref{app:search-resolution-sources}. The production pass never inspects document text, which keeps region attribution independent of the cultural-topic analysis in Section~\ref{sec:corpus-cultural-topic-annotation}.

\begin{table}[t]
\centering
\small
\setlength{\tabcolsep}{4pt}
\begin{tabular}{@{}llrr@{}}
\toprule
\textbf{Setting} & \textbf{Source\,/\,tier} & \textbf{Documents} & \textbf{\%} \\
\midrule
                     & \texttt{query}      & 15{,}283{,}520      & 0.05  \\
\addlinespace[2pt]
\shortstack[l]{lookup-table}
                     & \texttt{host\_path} & 222{,}919{,}273     & 0.72  \\
                     & \texttt{host}       & 3{,}458{,}247{,}367 & 11.19 \\
                     & \texttt{domain}     & 946{,}609{,}661     & 3.06  \\
\addlinespace[2pt]
fallback
                     & \texttt{url\_hint}  & 342{,}035{,}324     & 1.11  \\
                     & \texttt{tld}        & 2{,}932{,}873{,}160 & 9.49  \\

\midrule
confidence
                     & \texttt{high} (incl.\ \texttt{query})       & 2{,}384{,}519{,}175 & 7.71  \\
                     & \texttt{medium}     & 1{,}111{,}291{,}057 & 3.59  \\
                     & \texttt{low}        & 1{,}147{,}249{,}589 & 3.71  \\
\midrule
\multicolumn{2}{@{}l}{lookup-table coverage subtotal} & 4{,}627{,}776{,}301 & 14.97 \\
\multicolumn{2}{@{}l}{broad (any non-\texttt{XX})}    & \textbf{7{,}917{,}968{,}305} & 25.61 \\
\multicolumn{2}{@{}l}{unresolved (\texttt{none})}     & 22{,}996{,}190{,}454 & 74.39 \\
\midrule
\multicolumn{2}{@{}l}{\textbf{Total (FW+FW2)}}  & 30{,}914{,}158{,}759 & 100.00 \\
\bottomrule
\end{tabular}
\caption{Region-resolution diagnostics on the full FineWeb + FineWeb-2
release, by resolution source (top) and released confidence tier (middle). 
\texttt{high}+\texttt{medium}+\texttt{low} (15.02\%) equals lookup-table
coverage plus \texttt{query} (\texttt{high} includes \texttt{query});
the weak \texttt{url\_hint} and \texttt{ccTLD}
tiers map one-to-one onto the \texttt{url\_hint} and \texttt{tld} source rows
and are not repeated.
}
\label{tab:region-diagnostics}
\end{table}

\subsubsection{Confidence and Coverage Settings}
\label{sec:lookup-confidence}

Our lookup table contains 1{,}335{,}000 URL-signature rows. Confidence is assigned after evidence aggregation rather than per individual crawl record, and falls into three tiers (\texttt{high}, \texttt{medium}, \texttt{low}) combining vote count, vote share, and dump diversity; exact thresholds are given in Appendix~\ref{app:region-resolver-details}.

We report two coverage settings. The \textit{broad} setting retains any non-\texttt{XX} region emitted by the resolver, including weak URL-hint and ccTLD fallback labels. The \textit{lookup-table coverage} setting retains only labels whose resolution source is \texttt{host\_path}, \texttt{host}, or \texttt{domain}, i.e.\ labels backed by the frozen lookup table. \texttt{query} labels are high-confidence but not lookup-backed; they are excluded from lookup-table coverage by definition and reported separately (0.05\% of documents). This separation allows downstream users to trade off recall and precision according to the requirements of their analysis. Table~\ref{tab:region-diagnostics} reports the empirical distribution of resolution sources on the full release.

\subsubsection{Resolver Evaluation Against Independent References}
\label{sec:resolver-calibration}

Defining a single gold-standard region label for an arbitrary URL is difficult: the legal jurisdiction of the registrant, the physical hosting location, the content-declared locale, and the intended audience may all disagree (e.g., Egyptian-recipe content written in Arabic, hosted by a French CDN on a generic \texttt{.com} domain, with \texttt{og:locale=ar\_EG }). We therefore do not have a single reference standard for evaluation, and human annotations are impractical at this scale. Instead, we report \emph{convergence} between the resolver's output and two independent automatic reference signals, each capturing a different notion of ``region'':

\begin{table}[t]
\centering
\small
\begin{tabular}{lrrrr}
\toprule
\textbf{Tier} & \textbf{IP} & \textbf{og:locale} & \textbf{All\,$=$} & \textbf{All\,$\neq$} \\
\midrule
\texttt{high}        & 21.9 & 41.4 & 9.1  & 14.4 \\
\texttt{medium}      & 19.8 & 16.1 & 3.6  & 19.0 \\
\texttt{low}         & 17.1 &  9.6 & 3.8  & 22.6 \\
\texttt{broad-only}  & 37.8 & 17.1 & 9.3  & 33.2 \\
\bottomrule
\end{tabular}
\caption{Resolver evaluation against independent automatic references on stratified random samples of 1{,}000 URLs per tier, comparing three notions of ``region''.
Cells are percentages of the tier's URLs with the required signal(s) available. \textbf{All\,$=$}\,/\,\textbf{All\,$\neq$}: all three sources coincide\,/\,are pairwise distinct.}
\label{tab:resolver_convergence}
\end{table}

\begin{itemize}
    \item \textbf{IP geolocation} (MaxMind GeoLite2~\citep{maxmind_geolite2}), capturing physical hosting jurisdiction. For each
       URL we resolve the hostname to an IP address and look up its country in GeoLite2, yielding the jurisdiction of the server that serves the bytes. 
      
    \item \textbf{Content-declared locale}, captured from the HTML \texttt{lang} attribute~\citep{rfc5646} and the Open Graph
      \texttt{og:locale} property~\citep{ogp}. For each URL we re-fetch the page, parse the
      root \texttt{<html lang="…">} attribute and any \texttt{og:locale} meta tag, and take the region subtag (e.g.\ \texttt{en-GB} or
       \texttt{ar\_EG} $\to$ region \texttt{GB} or \texttt{EG}).

\end{itemize}

For each confidence tier of the resolver, Table~\ref{tab:resolver_convergence} reports the percentage of URLs whose label agrees with each reference, the percentage where the resolver and both references coincide, and the percentage where all three sources disagree --- which we interpret as URLs that are genuinely ambiguous rather than as resolver failures.

Two patterns are worth highlighting. First, the all-three-disagree fraction increases monotonically with decreasing resolver confidence (14.4\%, 19.0\%, 22.6\% across \texttt{high}, \texttt{medium}, \texttt{low}; rising to 33.2\% for \texttt{broad-only} URLs that have no lookup-backed evidence at all), indicating that the confidence tiers correctly identify which URLs are inherently ambiguous across reference signals. Second, agreement with \texttt{og:locale} drops sharply across tiers (41.4\% $\to$ 16.1\% $\to$ 9.6\%), confirming that high-tier URLs are also the ones whose content most frequently declares the same locale. Agreement with IP geolocation is broadly tier-invariant ($\sim$17--22\%) because IP measures physical hosting, which is largely orthogonal to the resolver's audit-unit semantics; the higher 37.8\% for \texttt{broad-only} reflects ccTLD- and hint-based URLs whose region label trivially aligns with hosting jurisdiction. Both reference signals are themselves imperfect: IP geolocation captures hosting rather than authorship, and \texttt{og:locale} is content-declared rather than verified. We therefore treat this analysis as a convergence study rather than a validation against ground truth, and recommend that downstream users rely on the released confidence tiers when interpreting individual region labels.

\subsection{Culture-Taxonomy Annotation}
  \label{sec:corpus-cultural-topic-annotation}

  The region axis introduced in Section~\ref{sec:region-attribution}
  identifies where a document plausibly originates from, but it does not
  identify what kind of cultural phenomenon is present in the document
  or its local corpus context. We therefore add cultural-topic labels
  to the region-resolved corpus. 

  \subsubsection{Taxonomy and Label Space}

\begin{figure}[t]
  \centering
  \resizebox{\columnwidth}{!}{%
  \begin{tikzpicture}[
    font=\sffamily\scriptsize,
    branch/.style={
      draw,
      rounded corners=3pt,
      line width=.55pt,
      align=left,
      text width=6.05cm,
      inner xsep=7pt,
      inner ysep=5pt
    },
    root/.style={
      draw=black!65,
      fill=black!4,
      rounded corners=2pt,
      line width=.55pt,
      font=\sffamily\bfseries\footnotesize,
      inner xsep=8pt,
      inner ysep=4pt
    },
    ide/.style={branch,draw=blue!60!black,fill=blue!3},
    lin/.style={branch,draw=green!45!black,fill=green!3},
    soc/.style={branch,draw=violet!60!black,fill=violet!3},
    connector/.style={draw=black!55,line width=.55pt},
    dot/.style={circle,inner sep=0pt,minimum size=3pt}
  ]

    \node[root] (root) {Cultural Taxonomy (CT)};

    \node[ide, below=3.5mm of root] (ide) {%
      {\bfseries\color{blue!60!black}Ideational \hfill 8 leaves}\\[1.5pt]
      \textbullet\ Concepts \qquad
      \textbullet\ Knowledge\\
      \textbullet\ Norms and Morals \qquad
      \textbullet\ Artifacts\\
      \textbullet\ Values $\rightarrow$ {\itshape General; Bias; Hate;}\\
      \hspace*{1.55em}{\itshape Other Perceptions}
    };

    \node[lin, below=1.2mm of ide] (lin) {%
      {\bfseries\color{green!45!black}Linguistic \hfill 2 leaves}\\[1.5pt]
      \textbullet\ Dialects \qquad
      \textbullet\ Styles, Registers and Genres
    };

    \node[soc, below=1.2mm of lin] (soc) {%
      {\bfseries\color{violet!60!black}Social \hfill 4 leaves}\\[1.5pt]
      \textbullet\ Relationship \qquad
      \textbullet\ Context\\
      \textbullet\ Communicative Goals \qquad
      \textbullet\ Demographics
    };

    \coordinate (spine) at ($(ide.west)+(-5mm,0)$);
    \draw[connector] (root.south) -- ++(0,-2.8mm) -| (spine);
    \draw[connector] (spine) |- (ide.west);
    \draw[connector] (spine) |- (lin.west);
    \draw[connector] (spine) |- (soc.west);

    \node[dot,fill=blue!60!black]   at (ide.west) {};
    \node[dot,fill=green!45!black]  at (lin.west) {};
    \node[dot,fill=violet!60!black] at (soc.west) {};


  \end{tikzpicture}%
  }
  \caption{Cultural Taxonomy label space used for corpus-side topic
  alignment. 
  }
  \label{fig:ct-label-space}
\end{figure}
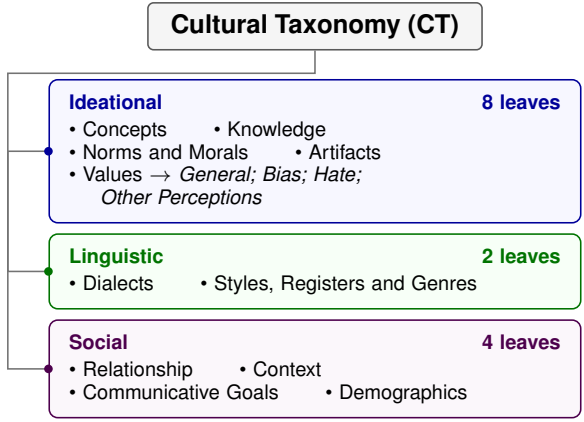

  We operationalize the Cultural Taxonomy (CT) of
  \citet{liu2025culturally} as the cultural label space. The
  CT organizes cultural phenomena into ideational, linguistic, and
  social dimensions. Figure~\ref{fig:ct-label-space} summarizes the
  label space used for topic alignment. The Values element of the
  Ideational branch is further refined into four sub-leaves, yielding $C=14$ leaf categories in total.
  The goal of our mapping is to estimate which CT
  categories are evidenced by a document's topic context.



\subsubsection{Locale-Specific Topic Induction}
\label{sec:locale-specific-topic-induction}

Human document-level CT annotation at the scale of FineWeb and FineWeb-2 is infeasible. We therefore use a topic labeling pipeline over language--region locales. Not all locales contain enough usable text for stable topic modeling: a locale is eligible when it contains at least 500 documents passing a quality filter (200--50{,}000 characters, language confidence $\geq 0.8$), which 6{,}332 locales meet. We sample up to 10{,}000 quality-filtered documents per locale; discarding locales whose retrieved sample falls below the 500-document threshold, yielding 5{,}954 topic-modeled locales spanning 297 languages (322 distinct language--script combinations). The resulting culturally annotated resource is therefore a subset of the full region-annotated corpus.

For each retained locale, we encode documents using multilingual BGE-M3 \citep{chen2024m3} and induce topics independently with FASTopic \citep{wu2024fastopic}. Modeling each locale separately prevents high-resource languages and globally dominant regions from determining the topic space for smaller locales, and allows locale-specific cultural content to form separate clusters. We use multilingual BGE-M3 because the pipeline requires a single shared embedding space in which document embeddings, topic centroids, and CT-leaf description embeddings are directly comparable by cosine similarity across all locales; we use its dense 1{,}024-dimensional embeddings. The language counts above describe the corpus language codes of the retained locales, not the validated language coverage of BGE-M3. Documents are truncated before encoding: the initial embedding pass uses BGE-M3's maximum input length of 8{,}192 tokens, and the consolidation pass that completed the final locale set caps inputs at 1{,}024 tokens.

We use a fixed initial maximum number of topics $K=200$ for every retained locale. This keeps the topic-induction hyperparameter constant across locales, which is important for the cross-locale comparisons in Section~\ref{sec:cross:locale_divergence}.\footnote{The fixed value should be interpreted as an overcomplete topic budget rather than as the final number of topics: the effective topic count is determined after the merge step.}

FASTopic \citep{wu2024fastopic} returns a document-topic distribution and topic descriptors. For each topic, we compute a centroid as the mean BGE-M3 embedding of the documents whose highest-probability topic assignment is that topic. We use these centroids for CT projection to the BGE-M3 CT leaf-description embeddings.

The fixed topic budget can produce near-duplicate topics, especially in smaller locales. We therefore merge topics using agglomerative average-linkage clustering from scikit-learn \citep{pedregosa2011scikit} over cosine distance between topic centroids, cutting clusters at cosine similarity $\geq 0.85$. For each merged topic, we sum the corresponding columns of the document-topic matrix, compute a document-count-weighted centroid, and retain the top unique words from the merged member topics. The merged topics are then passed to the topic-to-taxonomy projection step. Merging operates within each locale independently; across the 5{,}954 topic-modeled locales, the $K=200$ per-locale budget yields 801{,}344 merged topics in total, an effective average of $\approx$135 topics per locale.

  \subsubsection{Topic-to-Taxonomy Projection}

  We project topic centroids onto CT-leaf embeddings (both encoded
  with BGE-M3) via a temperature-scaled cosine softmax with $\tau = 0.05$;
  full equation and rationale in Appendix~\ref{app:methods}.

\subsubsection{Low-Confidence Adjudication}

The embedding-based projection is used as the default topic-to-CT mapping. Topics whose maximum CT probability falls below 0.5 receive LLM adjudication. Because the temperature-scaled softmax at $\tau = 0.05$ is diffused over the 14-leaf label space (observed $\max_c p(c \mid t)$: mean 0.124, 99th percentile 0.315), this criterion selects 99.97\% of merged topics (801{,}068 of 801{,}344); adjudication is therefore near-universal, and the 0.5 cutoff acts as a guardrail that bypasses adjudication only for unambiguous projections. The adjudicator receives the topic descriptors together with the top-three CT candidate categories and their softmax probabilities, and either confirms the automatic label or selects a better-fitting CT leaf; it changed the primary assignment for 7.1\% of adjudicated topics (56{,}931). Table~\ref{tab:example_topics} shows example FASTopic topics and their final CT leaf assignments for selected locales.

\begin{table}[h]
    \centering
    \small
    \setlength{\tabcolsep}{5pt}
    \renewcommand{\arraystretch}{1.05}
    \begin{tabular}{@{}lll@{}}
    \toprule
    \textbf{Locale} & \textbf{Topic} & \textbf{CT leaf} \\
    \midrule
    en-US & Christian Religion & Values-General \\
    en-US & Restaurant Dining & Concepts \\
    en-US & US Politics & Demographics \\
    ar-EG & Travel Destinations & Concepts \\
    ar-EG & Sports Media & Context \\
    zh-CN & Digital Media \& Devices & Artifacts \\
    hi-IN & Indian Culture & Knowledge \\
    ko-KR & Korean Global Education & Demographics \\
    \bottomrule
    \end{tabular}
    \caption{Example FASTopic topics and final CT leaf assignments for selected locales.}
    \label{tab:example_topics}
\end{table}

\subsubsection{Locale Topic Distributions}

  The output of the corpus-side cultural-topic pipeline is a Locale Topic Distributions (LTD) for each topic-modeled locale. For a locale
  with $N$ documents, $K$ merged topics, and $C$ CT categories, let
  $D \in \mathbb{R}^{N \times K}$ be the document--topic distribution
  and $A \in \mathbb{R}^{K \times C}$ be the topic-to-CT assignment
  matrix. The locale's Locale Topic Distribution is:
  \[
  \mathrm{LTD}
  =
  \frac{1}{N}
  \sum_{i=1}^{N}
  (DA)_i.
  \]
  The resulting vector is L1-normalized and interpreted as the locale's
distribution over CT leaves. Across all topic-modeled locales,
this procedure produces a $5{,}954 \times C$ matrix of LTD, with $C=14$. We use this matrix for distributional
analyses of locale-level cultural-topic profiles, including the
same-language regional divergence analysis in
Section~\ref{sec:cross:locale_divergence}.

\section{Benchmark Audit}
\label{sec:benchmark-audit}

To compare corpus-side cultural evidence with evaluation demand, we
construct a benchmark-side audit using the same Cultural Taxonomy label space and
locale metadata as FineWeb-CLaR. The audit extends prior
taxonomy-driven cultural-NLP surveys \citep{liu2025culturally,namazifard2025cultureneurons} by adding newly identified
datasets and by coding each resource for cultural, language, and region coverage, provenance, and usability.

\begin{figure}[t]
    \centering
\definecolor{clsNeither}{HTML}{F0F0F0}
\definecolor{clsCorpus}{HTML}{56B4E9}
\definecolor{clsSurvey}{HTML}{E69F00}
\definecolor{clsBoth}{HTML}{004488}
\definecolor{OIblue}{HTML}{0072B2}
\definecolor{medblue}{HTML}{1B6CA8}
\definecolor{lightblue}{HTML}{A6CBE3}
\definecolor{oiorange}{HTML}{D95F02}
\definecolor{histblue}{HTML}{4A87E0}
\definecolor{darkblue}{HTML}{123E85}
\begin{tikzpicture}
\begin{axis}[
  width={\dimexpr\linewidth-42pt\relax}, height=0.74\linewidth,
  xbar, bar width=4.4pt,
  xmin=0, xmax=40, ymin=0.4, ymax=14.6,
  xtick={0,5,...,40},
  ytick={1,...,14},
  yticklabels={{Relationship},{Communicative Goals},{Values - other},{Artifacts},{Demographics},{Values - hate},{Values - general},{Dialects},{Styles/Reg./Genres},{Context},{Concepts},{Values - bias},{Knowledge},{Norms and Morals}},
  axis x line*=bottom, axis y line*=left,
  xlabel={number of datasets},
  tick label style={font=\scriptsize},
  label style={font=\scriptsize},
  ytick style={draw=none},
  nodes near coords={\pgfmathprintnumber[fixed, precision=0]{\pgfplotspointmeta}},
  every node near coord/.append style={font=\scriptsize, xshift=1pt},
]
\addplot[xbar, fill=OIblue, draw=none] coordinates { (7,1) (11,2) (12,3) (12,4) (12,5) (13,6) (15,7) (15,8) (16,9) (17,10) (21,11) (28,12) (32,13) (36,14) };
\end{axis}
\end{tikzpicture}
    \caption{Number of audited benchmark datasets per CT leaf ($n=277$). 
    }
    
    \label{fig:survey-original-vs-expansion-category}
\end{figure}
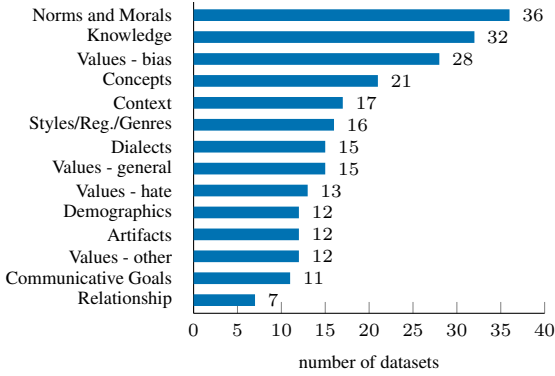

\paragraph{Dataset collection.}
We compile cultural NLP benchmarks from three sources: datasets included
in the cultural-NLP taxonomy and survey of \citet{liu2025culturally},
datasets included in MUREL-style cultural-resource curation
\citep{namazifard2025cultureneurons}, and newly identified resources from
systematic searches of the ACL Anthology, arXiv, HuggingFace, GitHub,
and dataset repositories. After deduplication and filtering, the audit
contains 277 cultural NLP benchmark datasets (43 inherited from the Liu et al.\ taxonomy survey, 3 from MUREL, 19 shared by both, and 212 newly identified). A resource is included
when it provides an accessible dataset, targets at least one cultural
phenomenon covered by the CT, and is documented well enough to recover
its language, region, task, and provenance metadata. Figure~\ref{fig:survey-original-vs-expansion-category} shows the per-leaf coverage of the expanded audit. The newly identified datasets
increase coverage across most CT leaves, including linguistic and social
categories that are comparatively sparse in the inherited survey sources.

\paragraph{Annotation schema.}
Each benchmark is annotated along four groups of fields. Cultural
coverage records the CT branch and one or more CT leaves targeted by
the dataset. Scope records the covered languages, regions and example items when available. Provenance records the source paper,
venue, publication date, source list, and dataset
lineage. Usability records whether the dataset is public, restricted,
or unavailable.

\begin{figure*}[t]
    \centering
    \input{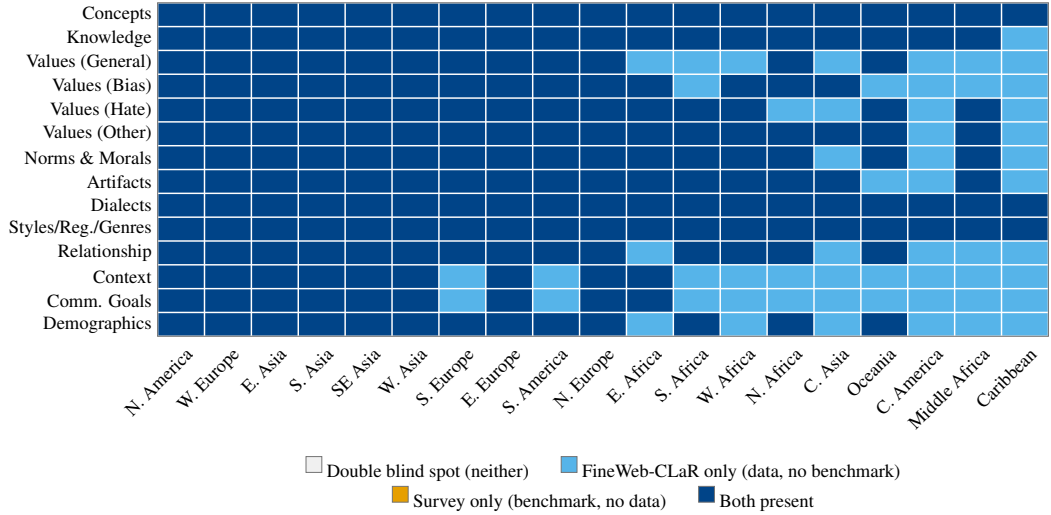}
        \caption{Coverage over CT leaf $\times$ region cells on the full release (30.9B documents). A cell is corpus-covered if at least $k=50$ documents support the CT-leaf/region pair, and benchmark-covered if at least one audited benchmark targets it. All 266 cells are corpus-covered; light cells (56. 21\%) have corpus evidence but no audited benchmark.}

    \label{fig:blind_spots}
\end{figure*}

\paragraph{Human annotation and validation.} Two authors annotated the benchmarks and resolved disagreements through discussion. CT branch and CT leaf labels were assigned using each dataset's paper, dataset card, repository, and example items. A dataset may receive multiple CT leaves when it evaluates more than one cultural phenomenon. Details of the search protocol, annotation rubric, lineage coding, accessibility coding, and quality-control procedure are provided in Appendix~\ref{app:benchmark-audit}. Inter-annotator agreement on the multi-label Cultural Taxonomy assignments was 0.43 average pairwise Jaccard, with 0.89 partial agreement when any shared CT branch/leaf label counted as overlap.\footnote{We release the full benchmark survey in our code repository.}

\section{Experiments and Analysis}
\label{sec:experiments}

We evaluate FineWeb-CLaR as a corpus-auditing resource through three analyses: human annotation, corpus-benchmark mismatch, and same-language regional divergence.



\subsection{Validation of the Topic-to-CT Projection}
\label{sec:topic-ct-validation}

Two annotators independently coded 100 (topic, CT-leaf) pairs sampled
stratified across the 14 CT leaves and across high-, medium-, and
low-resource locales. Each pair was presented with the top-15 FASTopic
descriptors, ten randomly drawn documents from the topic, and the
natural-language definition of each CT leaf. Annotators selected the
single best-fitting CT leaf or \texttt{NONE}. Annotators did not see the
embedding-softmax or LLM-adjudicated labels during annotation.

Cohen's $\kappa$ between annotators was 0.25, indicating only fair
agreement on the topic-to-CT-leaf task and confirming that CT-leaf
assignment at the topic level is intrinsically ambiguous. On 15\% of
topics the two annotators could not converge on a single leaf; the
remaining $N=85$ topics form the consensus reference set. Against this
reference, the embedding-softmax label agreed on 18.8\% of cases
(Cohen's $\kappa$ = 0.13); the LLM adjudicator, which was triggered on
all sampled topics and altered the embedding-softmax assignment for
35\% of them, agreed on 22.4\% of cases (Cohen's $\kappa$ = 0.15). 
Both automatic methods are well below human consensus, and the
adjudicator yields only marginal improvements over the embedding
baseline. We therefore interpret LTDs not as definitive CT-leaf
assignments but as locale-level distributions over topic-to-leaf
\emph{prevalences} under substantial labeling uncertainty.
Accordingly, we treat the CT-level analyses that follow as exploratory corpus audits: the coverage classification in Section~\ref{sec:cross:gaps} depends on document counts per cell rather than on individual leaf assignments, and the divergence analysis in Section~\ref{sec:cross:locale_divergence} compares whole distributions between locales that share a language, but both inherit the label uncertainty quantified here. Appendix~\ref{app:robustness} quantifies this inheritance by resampling every topic-to-leaf assignment at the observed agreement rate ($B = 1000$ replicates): the count-based coverage results and the pairs-above-noise-floor finding hold in every replicate of every condition, while exact pair orderings are only partially preserved. 



\subsection{Corpus--Benchmark Mismatch}
\label{sec:cross:gaps}

The shared language--region--CT axis allows us to compare corpus evidence with benchmark demand. On the corpus side, FineWeb-CLaR provides document counts and Locale Topic Distributions for language--region locales. On the benchmark side, the human annotation in Section~\ref{sec:benchmark-audit} provides CT leaves, language coverage, and region coverage for 277 cultural NLP benchmarks.

We mark a CT leaf--region cell as corpus-covered when at least $k=50$ documents are associated with that cell. Each cell is then classified as both present, corpus only, benchmark only, or neither present, depending on whether it appears in FineWeb-CLaR, the benchmark inventory, both, or neither. 

Figure~\ref{fig:blind_spots} shows that, on the full release, corpus evidence exceeds the $k=50$ threshold for all 266 CT-leaf $\times$ region cells, so the mismatch is benchmark-side: 56 cells (21\%) carry corpus evidence but no audited benchmark, concentrated regionally in the Caribbean (11 of 14 leaves), Central America (10), Central Asia (7), and Middle Africa (6), and thematically in the Social branch (Context 10, Communicative Goals 10) and the Values leaves (18). Ideational categories such as \emph{Knowledge}, \emph{Concepts}, and \emph{Artifacts} are more often visible in web text and more often converted into benchmark tasks. Thus, corpus--benchmark mismatch is not only a language-level problem; it also depends on which cultural phenomenon is being evaluated.

\begin{figure*}[t]
    \centering
    \input{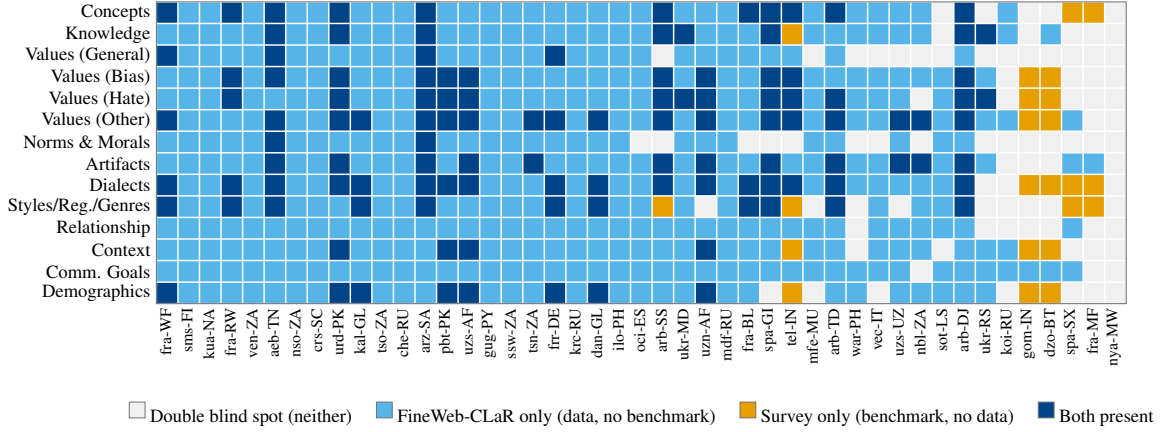}
    \caption{CT leaf $\times$ locale coverage for the smallest
    topic-modeled main-language locales on the full release.
    Locale-level analysis reveals the double blind spots and
    benchmark-only cells that macro-region aggregation hides.}
    \label{fig:locale_matrix}
\end{figure*}

Earlier versions of this analysis, computed on analysis corpora in which English was represented through the FineWeb \texttt{sample-350BT} subset, showed several region-level double blind spots, including the entire Caribbean column. Recomputing the analysis on the full release provides a more complete picture: these patterns do not persist at full scale and were primarily driven by the $\sim$50$\times$ downsampling of English rather than by an absence of corpus evidence (see Appendix~\ref{app:datasheet} for details on the different sampling experiments). The change is specific to English: English-locale mass in these regions increases by 46--56$\times$ on the full release, while their non-English main-language mass remains unchanged. The full-release analysis therefore resolves the apparent macro-region blind spots observed in earlier versions while revealing more fine-grained coverage differences at the locale level (Figure~\ref{fig:locale_matrix}). Applying the same classification at locale granularity shows that 26.5\% of cells are covered by both corpus and benchmarks, 63.9\% by the corpus only, 1.9\% by benchmarks only, and 7.7\% by neither. Chichewa (nya-MW; 546 documents) lacks both corpus mass and benchmarks across all 14 leaves; Dzongkha (dzo-BT) is benchmark-covered but corpus-sparse on 6 leaves; and even en-GB, the largest locale (1.33B documents), has no benchmark coverage on 8 of 14 leaves. Overall, 863 locales (14.8\%) are covered by no region-scoped benchmark, including 67 main-language locales in their respective countries (e.g., Bosnian bs-BA with 3.0M documents). Figure~\ref{fig:locale_matrix} shows the locale-level coverage matrix for the smallest main-language locales.


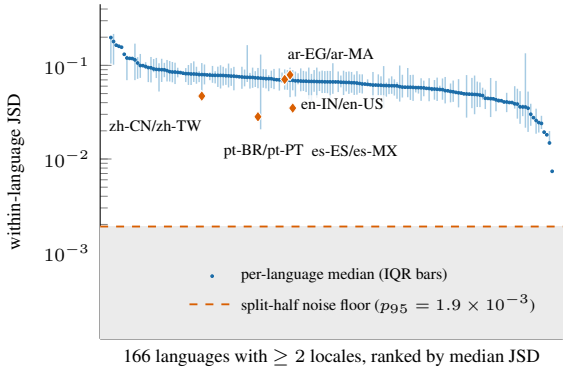
\begin{figure}[t]
    \centering
\definecolor{clsNeither}{HTML}{F0F0F0}
\definecolor{clsCorpus}{HTML}{56B4E9}
\definecolor{clsSurvey}{HTML}{E69F00}
\definecolor{clsBoth}{HTML}{004488}
\definecolor{OIblue}{HTML}{0072B2}
\definecolor{medblue}{HTML}{1B6CA8}
\definecolor{lightblue}{HTML}{A6CBE3}
\definecolor{oiorange}{HTML}{D95F02}
\definecolor{histblue}{HTML}{4A87E0}
\definecolor{darkblue}{HTML}{123E85}
\begin{tikzpicture}
\begin{axis}[
  width=\linewidth, height=0.78\linewidth,
  ymode=log, ymin=1.2e-4, ymax=0.45, xmin=-4, xmax=169,
  xtick=\empty,
  axis x line*=bottom, axis y line*=left,
  ylabel={within-language JSD},
  xlabel={166 languages with $\geq 2$ locales, ranked by median JSD},
  label style={font=\scriptsize},
  tick label style={font=\scriptsize},
  legend style={at={(0.97,0.04)}, anchor=south east, draw=none, fill=none, font=\tiny},
  legend cell align=left,
]
\fill[black!8] (axis cs:-4,1.2e-4) rectangle (axis cs:169,0.00190);
\addplot[medblue, only marks, mark=*, mark size=0.5pt,
  error bars/.cd, y dir=both, y explicit,
  error bar style={lightblue, line width=0.55pt}, error mark=none,
] table[x index=0, y index=1, y error minus index=2, y error plus index=3] {
0 0.19850 0.09389 0.00059
1 0.17995 0.07828 0.03675
2 0.16440 0.00000 0.00000
3 0.16005 0.00000 0.00000
4 0.15637 0.00000 0.00000
5 0.13141 0.00000 0.00000
6 0.11992 0.05042 0.00701
7 0.11883 0.00000 0.00000
8 0.11835 0.00000 0.00000
9 0.11453 0.04496 0.05620
10 0.10602 0.03418 0.01645
11 0.09964 0.02606 0.03281
12 0.09939 0.00000 0.00000
13 0.09725 0.00585 0.01094
14 0.09481 0.02849 0.00300
15 0.09424 0.02852 0.02874
16 0.09159 0.00695 0.02530
17 0.09066 0.01736 0.02949
18 0.09018 0.03758 0.01523
19 0.08993 0.02342 0.02772
20 0.08964 0.01478 0.01676
21 0.08790 0.02056 0.02632
22 0.08558 0.01833 0.02568
23 0.08499 0.01904 0.02372
24 0.08489 0.02169 0.02189
25 0.08427 0.00974 0.01009
26 0.08375 0.00000 0.00000
27 0.08251 0.00000 0.00000
28 0.08159 0.01523 0.01199
29 0.08152 0.01968 0.01945
30 0.08097 0.01924 0.02475
31 0.08094 0.01448 0.02944
32 0.08062 0.01697 0.01607
33 0.07976 0.01351 0.00689
34 0.07975 0.02721 0.02616
35 0.07957 0.01692 0.02265
36 0.07909 0.00000 0.00000
37 0.07881 0.00931 0.00507
38 0.07837 0.01321 0.02029
39 0.07832 0.01408 0.00841
40 0.07821 0.01941 0.01873
41 0.07818 0.02154 0.02218
42 0.07807 0.01716 0.01856
43 0.07755 0.01561 0.01553
44 0.07735 0.01918 0.02723
45 0.07684 0.01641 0.01639
46 0.07678 0.02260 0.01969
47 0.07617 0.00000 0.00000
48 0.07562 0.00000 0.00000
49 0.07518 0.01897 0.02150
50 0.07478 0.01504 0.01548
51 0.07462 0.00812 0.08993
52 0.07444 0.01794 0.00871
53 0.07438 0.01683 0.02026
54 0.07368 0.01489 0.01312
55 0.07344 0.01641 0.02047
56 0.07303 0.05227 0.05751
57 0.07269 0.01497 0.01853
58 0.07213 0.01534 0.01873
59 0.07209 0.01905 0.00651
60 0.07205 0.01484 0.01577
61 0.07090 0.01511 0.02079
62 0.07025 0.00000 0.00000
63 0.07005 0.01625 0.02282
64 0.06977 0.00489 0.01409
65 0.06976 0.01469 0.02075
66 0.06956 0.01475 0.01948
67 0.06936 0.02358 0.02515
68 0.06918 0.01499 0.01980
69 0.06840 0.02187 0.01755
70 0.06835 0.00553 0.01401
71 0.06797 0.01507 0.02493
72 0.06778 0.01443 0.01963
73 0.06776 0.01585 0.01833
74 0.06744 0.02378 0.02733
75 0.06738 0.02580 0.01374
76 0.06736 0.01509 0.02178
77 0.06713 0.02202 0.02757
78 0.06698 0.01375 0.01908
79 0.06696 0.01630 0.02100
80 0.06672 0.01601 0.02019
81 0.06669 0.01625 0.02070
82 0.06665 0.02176 0.02055
83 0.06665 0.01694 0.03887
84 0.06659 0.00000 0.00000
85 0.06656 0.01468 0.01956
86 0.06610 0.00781 0.01182
87 0.06599 0.01781 0.02172
88 0.06574 0.01294 0.01951
89 0.06574 0.01614 0.02045
90 0.06526 0.01422 0.01980
91 0.06524 0.01691 0.01448
92 0.06476 0.01608 0.02201
93 0.06410 0.01323 0.01618
94 0.06333 0.01323 0.01692
95 0.06316 0.00000 0.00000
96 0.06266 0.00250 0.00837
97 0.06218 0.00000 0.00000
98 0.06166 0.02547 0.02336
99 0.06165 0.00000 0.00000
100 0.06144 0.01355 0.01874
101 0.06135 0.01351 0.01509
102 0.06095 0.02601 0.03735
103 0.06084 0.01452 0.00465
104 0.06066 0.01134 0.01186
105 0.06060 0.01091 0.01574
106 0.06055 0.01292 0.01044
107 0.06047 0.01648 0.00566
108 0.05929 0.01821 0.04035
109 0.05866 0.01329 0.02074
110 0.05848 0.00000 0.00000
111 0.05843 0.01273 0.03534
112 0.05842 0.01566 0.02058
113 0.05827 0.01652 0.02114
114 0.05790 0.00646 0.00027
115 0.05783 0.00782 0.00287
116 0.05767 0.01586 0.01522
117 0.05750 0.01069 0.01126
118 0.05685 0.00000 0.00000
119 0.05667 0.01528 0.01632
120 0.05594 0.00336 0.00305
121 0.05588 0.01285 0.01391
122 0.05563 0.01395 0.01841
123 0.05521 0.00025 0.00544
124 0.05511 0.01215 0.01739
125 0.05375 0.01450 0.02147
126 0.05353 0.00000 0.00000
127 0.05265 0.01236 0.01142
128 0.05226 0.01148 0.01662
129 0.05148 0.00000 0.00000
130 0.05094 0.01533 0.02033
131 0.05058 0.00000 0.00000
132 0.04970 0.00000 0.00000
133 0.04949 0.02181 0.02930
134 0.04929 0.00000 0.00000
135 0.04914 0.01422 0.02437
136 0.04892 0.01884 0.02897
137 0.04874 0.01282 0.01838
138 0.04819 0.01235 0.01753
139 0.04762 0.00456 0.01013
140 0.04456 0.01038 0.01046
141 0.04442 0.00000 0.00000
142 0.04409 0.01602 0.02925
143 0.04395 0.00000 0.00000
144 0.04289 0.01049 0.02544
145 0.04195 0.00078 0.00268
146 0.04193 0.00000 0.00000
147 0.04068 0.00000 0.00000
148 0.04066 0.00899 0.01027
149 0.04017 0.00000 0.00000
150 0.04011 0.00000 0.00000
151 0.03886 0.00432 0.02107
152 0.03862 0.01163 0.01768
153 0.03609 0.01233 0.01697
154 0.03598 0.00728 0.00219
155 0.03596 0.01436 0.09867
156 0.03487 0.01540 0.01765
157 0.03006 0.00742 0.00887
158 0.02766 0.00000 0.00000
159 0.02579 0.00716 0.00869
160 0.02451 0.00533 0.00020
161 0.02400 0.00307 0.00493
162 0.01936 0.00000 0.00000
163 0.01815 0.00000 0.00000
164 0.01484 0.00050 0.00505
165 0.00739 0.00000 0.00000
};
\addlegendentry{per-language median (IQR bars)}
\addplot[oiorange, dashed, line width=0.7pt] coordinates {(-4,0.00190) (169,0.00190)};
\addlegendentry{split-half noise floor ($p_{95} = 1.9 \times 10^{-3}$)}
\addplot[only marks, mark=diamond*, oiorange, mark size=2.1pt,
  mark options={draw=white, line width=0.3pt}, forget plot] coordinates { (34,0.04702) (55,0.02826) (65,0.07083) (67,0.07918) (68,0.03499) };
\node[font=\tiny, fill=white, fill opacity=0.78, text opacity=1, inner sep=1pt, rounded corners=1pt, xshift=-36pt, yshift=-11pt, anchor=west] at (axis cs:34,0.04702) {zh-CN/zh-TW};
\node[font=\tiny, fill=white, fill opacity=0.78, text opacity=1, inner sep=1pt, rounded corners=1pt, xshift=-14pt, yshift=-13pt, anchor=west] at (axis cs:55,0.02826) {pt-BR/pt-PT};
\node[font=\tiny, fill=white, fill opacity=0.78, text opacity=1, inner sep=1pt, rounded corners=1pt, xshift=5pt, yshift=-9pt, anchor=west] at (axis cs:65,0.07083) {en-IN/en-US};
\node[font=\tiny, fill=white, fill opacity=0.78, text opacity=1, inner sep=1pt, rounded corners=1pt, xshift=-2pt, yshift=8pt, anchor=west] at (axis cs:67,0.07918) {ar-EG/ar-MA};
\node[font=\tiny, fill=white, fill opacity=0.78, text opacity=1, inner sep=1pt, rounded corners=1pt, xshift=6pt, yshift=-16pt, anchor=west] at (axis cs:68,0.03499) {es-ES/es-MX};
\end{axis}
\end{tikzpicture}
    \caption{Within-language LTD divergence (JSD) across all languages with $\geq 2$ locales: per-language median (dots) and IQR (bars), with illustrative same-language locale pairs labelled. The shaded band and dashed line show the split-half noise floor ($p_{95} = 1.9 \times 10^{-3}$).}
    \label{fig:within_language_jsd}
\end{figure}

\subsection{Same-Language Regional Divergence}
\label{sec:cross:locale_divergence}
Language-level corpus partitions can hide regional variation. We quantify this by computing the Jensen--Shannon distance (JSD full definition in Appendix~\ref{app:methods}) between Locale Topic Distributions for locales that share a language but differ in region. Higher JSD indicates greater divergence between the two locale-level CT distributions.

To distinguish genuine regional divergence from topic-modeling noise, we compute a within-locale split-half JSD baseline. For each of the 4{,}715 locales with at least 1{,}000 sampled documents, we randomly partition the locale's documents into two halves, fit an LTD on each half, and compute the JSD between halves; we repeat this 50 times per locale (235{,}750 split-half computations in total, seed 42). The pooled distribution yields a median JSD of $4.9 \times 10^{-4}$ (IQR $[2.4 \times 10^{-4},\, 8.8 \times 10^{-4}]$, $p_{95} = 1.9 \times 10^{-3}$), which we use as the topic-model noise floor. The floor decreases with locale size (full details in Appendix~\ref{app:robustness-floor}); the five labelled pairs below involve locales at the 10{,}000-document cap, where the $p_{95}$ floor is roughly half the pooled value, so the pooled floor if anything understates their separation from noise.

Figure~\ref{fig:within_language_jsd} shows the distribution of within-language JSD across every language with at least two locales meeting the 500-document floor. The labelled examples all lie well above the split-half noise floor, indicating that same-language regional differences in LTDs are substantially larger than topic-modeling noise. The largest labelled divergences occur for the Arabic and English locale pairs (ar-EG/ar-MA 0.079, en-IN/en-US 0.071), followed by Chinese (zh-CN/zh-TW 0.047), while the Spanish and Portuguese pairs (0.035, 0.028) show smaller but still clearly non-noise divergence. We interpret this as evidence that benchmarks targeting a language may correspond to specific regional CT profiles, while noting that the magnitude of the language-level/regional gap varies substantially across language families.

\section{Conclusion}
We introduced FineWeb-CLaR, a large-scale resource that augments FineWeb and FineWeb-2 with region metadata and Cultural Taxonomy-aligned topic annotations to support culture-, language-, and region-aware corpus auditing. By combining URL-derived region attribution, locale-specific topic modeling, and an expanded audit of cultural NLP benchmarks, FineWeb-CLaR enables direct comparison between pretraining evidence and evaluation demand. Our analyses show that pretraining evidence at web scale exists for every cultural-topic $\times$ region cell we audit, but is unevenly distributed across regions by more than two orders of magnitude, while benchmark coverage is absent for 21\% of region-level cells; at locale granularity, coverage gaps run in both directions, with 7.7\% of locale $\times$ cultural-topic cells lacking both corpus evidence and benchmarks. Corpus--benchmark mismatch is therefore structured by region and cultural phenomenon, and must be audited at locale level, beyond language-only metadata.

\section*{Limitations}
\label{sec:limitations}

FineWeb-CLaR is an auditing resource, not demographic or cultural ground truth. Its language, region, and CT axes make corpora and benchmarks comparable, but they do not define cultures or communities. Region labels are derived from URL evidence and should not be interpreted as author location, user location, national identity, or cultural membership. They may be affected by multi-region websites, generic or vanity domains, CDNs, URL parsing errors, branded ccTLDs, and the use of a sampled Common Crawl lookup table. False positives and false negatives are therefore unavoidable.

The CT annotation pipeline is also coarse. CT labels support corpus-level comparison, but they cannot capture the full complexity of cultural practice, identity, power, tacit knowledge, interactional norms, or intra-community variation. Locale-specific topics may reflect domain composition, crawl artefacts, or institutional web presence rather than cultural salience alone. Embedding-based CT ranking and LLM-assisted adjudication reduce annotation cost, but can introduce errors, especially for low-resource languages, mixed-language documents, and culturally specific concepts that are weakly represented in the embedding model or adjudicator. Human projection quality is validated only in aggregate (Section~\ref{sec:topic-ct-validation}), and BGE-M3's validated language coverage is narrower than the corpus language inventory, so CT-level results should be read as exploratory; computationally, 67\% of merged topics in locales outside BGE-M3's documented coverage project onto the taxonomy with near-uniform confidence, against 46\% inside (Appendix~\ref{app:robustness-bgem3}), so LTDs for out-of-coverage languages carry additional uncertainty. Empirically, locale evidence is often domain-concentrated: the median topic-modeled locale draws on 116 distinct base domains, and 34\% of locales draw more than half of their documents from a single domain (Figure~\ref{fig:url_domain_diversity}); 83 locales (1.4\%) rest on a single base domain (verified exhaustively against the full release), typically a multilingual portal or machine-translated content farm attributed to one region. Locale-level distributions for such locales should be read as summaries of a few dominant sites rather than of a broad regional web.

Locale Topic Distributions are available only for locales with sufficient document support. Locales below the 500 quality-filtered-document threshold remain region-annotated, but do not receive LTDs because their topic distributions would be dominated by sampling noise. Even among the 5{,}954 topic-modeled locales, smaller locales should be interpreted with greater uncertainty.

The benchmark survey is time-bounded and depends on the availability and documentation of existing datasets. Human CT coding reduces noise, but does not establish that a benchmark validly represents the community or cultural phenomenon it claims to test. The coverage matrix should therefore be read as a diagnostic of available resources, not as a normative statement about which cultures or cultural dimensions matter.

\section*{Ethics Statement}
\label{sec:ethics}

This work annotates public web-corpus metadata for auditing and analysis. Because region labels can be misused if treated as demographic labels, we release confidence and source metadata and recommend against using weak labels for claims about people, identity, or community representation. Cultural-taxonomy annotations should likewise be used to identify broad coverage patterns, not to essentialise cultures or reduce communities to fixed attributes. Downstream use for data filtering, benchmark construction, or model evaluation should document the selected confidence threshold, the intended interpretation of region labels, and the limitations of URL-derived evidence.

\section*{Acknowledgments}
This research was supported by the German Federal Ministry for Economic Affairs and Energy (BMWE) as part of the project \textit{“Souveräne KI für Europa (SOOFI)”} (13IPC040H), and by the German Federal Ministry of Research, Technology and Space (BMFTR) as part of the project TRAILS (01IW24005). We thank Julia Kreutzer for her valuable feedback on an early draft of the paper. We also thank AriaRay Brown for helping with a part of the human annotations on the topic assignment. 

\bibliography{custom,datasets}

\appendix

\section{Region Resolver Details}
\label{app:region-resolver-details}

During lookup construction, each Common Crawl sample record is assigned a weighted vote according to the agreement pattern among URL evidence, HTTP \texttt{Content-Language}, HTML \texttt{lang}, and language-identification metadata. The rules are used only during offline lookup construction; they do not fire during FineWeb/FineWeb-2 annotation.

\paragraph{Confidence-tier thresholds.}
After evidence aggregation, each URL-signature row is assigned a confidence tier based on the following thresholds. A row is labelled \texttt{high} if the winning region has at least 20 weighted votes, at least 90\% vote share, and appears across at least three Common Crawl dumps. It is labelled \texttt{medium} if it has at least seven weighted votes, at least 75\% vote share, and appears across at least two dumps. It is labelled \texttt{low} if it has at least three weighted votes and at least 70\% vote share. These thresholds favour URL signatures that are stable across time and not merely frequent in a single crawl; we selected them empirically by manually inspecting samples of candidate lookup rows at different cutoff values before freezing the table, and no downstream corpus statistics or analyses were consulted during threshold selection.

\paragraph{Real ccTLD evidence.}
When the URL contains a non-branded country-code top-level domain, the ccTLD supplies the primary region candidate. Three-way agreement among ccTLD, HTML \texttt{lang}, and HTTP \texttt{Content-Language} receives the highest weight. Agreement between the ccTLD and one header signal receives slightly lower weight. When the ccTLD is contradicted by a likely CDN-default header, the ccTLD is still trusted but downweighted. When the ccTLD is contradicted by a country-specific header, the vote is retained with weaker weight. If no header signal is available, the ccTLD alone still contributes a strong but not maximal vote.

\paragraph{Path and subdomain evidence.}
For generic or branded domains, the system searches for locale-bearing path or subdomain tokens, such as \texttt{/uk/}, \texttt{/jp/}, \texttt{de.example.com}, or \texttt{en-gb}. These hints receive higher weight when corroborated by HTTP or HTML language metadata and lower weight when they appear alone.

\paragraph{Header-only evidence.}
If the URL contains no region cue, low-resolution header evidence may contribute a weak vote only when it is consistent with language identification. High-resolution headers on otherwise generic URLs are not treated as positive region evidence because such values are often injected by server or CDN defaults.

\paragraph{Fallback and unresolved records.}
If no reliable rule supplies evidence during lookup construction, the record contributes no vote. During search-time resolution, URLs that do not match the lookup table may still receive weak labels from curated URL hints or non-branded ccTLD fallback; otherwise the resolver emits \texttt{XX}.

\paragraph{Per-rule vote weights.}
Each rule contributes a per-record vote with a fixed weight, calibrated to reflect the strength of evidence it provides. The weights used during offline lookup construction are summarized in Table~\ref{tab:per-rule-weights}.

\begin{table*}[h]
\centering
\begin{tabular}{@{}lll@{}}
\toprule
\textbf{Rule} & \textbf{Evidence pattern} & \textbf{Weight} \\
\midrule
A1 & ccTLD + HTML \texttt{lang} + Content-Language agree  & 3.0 \\
A2 & ccTLD + one header agree                             & 2.5 \\
A3 & ccTLD beats high-resolution default                  & 2.0 \\
A4 & ccTLD overrides country-specific header              & 1.0 \\
A5 & ccTLD alone (no header signal)                       & 2.0 \\
A6 & Path/subdomain hint + header                         & 2.0 \\
A7 & Path/subdomain hint alone                            & 1.5 \\
A8 & Low-resolution header + language ID agree            & 1.5 \\
A9 & Soft header default on generic URL                   & 0  \\
A10 & Unresolved                                          & 0 \\
\bottomrule
\end{tabular}
\caption{Vote weights applied by each evidence rule during offline lookup-table construction. Higher weights correspond to stronger multi-signal agreement. Rules A9 and A10 are recorded for diagnostic completeness but do not contribute to weighted aggregation.}
\label{tab:per-rule-weights}
\end{table*}

\paragraph{Lookup-Table Schema}
The frozen URL-signature lookup table is released as a CSV with 1{,}335{,}000 rows and the columns described in Table~\ref{tab:resolver-schema}.

\begin{table*}[h]
\centering
\setlength{\tabcolsep}{4pt}
\begin{tabular}{@{}llp{8.5cm}@{}}
\toprule
\textbf{Column} & \textbf{Dtype} & \textbf{Meaning / values} \\
\midrule
\texttt{key\_type}            & str   & \texttt{domain}, \texttt{host}, or \texttt{host\_path} \\
\texttt{key}                  & str   & URL signature (normalised, lowercased) \\
\texttt{n\_docs}              & int   & raw record count contributing to the signature \\
\texttt{top\_region}          & str   & winning region (ISO~3166-1 alpha-2) or empty \\
\texttt{share}                & float & vote share of the winning region $\in[0,1]$ \\
\texttt{n\_distinct\_regions} & int   & distinct candidate regions seen \\
\texttt{confidence}           & str   & \texttt{high}, \texttt{medium}, or \texttt{low} \\
\texttt{n\_weighted}          & float & total weighted votes for the winning region \\
\texttt{dump\_diversity}      & int   & number of distinct CC-MAIN dumps the row appears in \\
\texttt{secondary\_region}    & str   & runner-up region or empty \\
\texttt{secondary\_share}     & float & vote share of the runner-up \\
\texttt{n\_content\_votes}    & int   & A11 content-scan votes (0 in URL-only release) \\
\texttt{content\_share}       & float & A11 content-share $\in[0,1]$ (0 in URL-only release) \\
\texttt{top\_rule}            & str   & rule that contributed the largest mass (\texttt{A1}--\texttt{A12}) \\
\texttt{rule\_mix}            & str   & semicolon-separated \texttt{Ak:count} list across all firing rules \\
\bottomrule
\end{tabular}
\caption{Schema of the released URL-signature lookup table.}
\label{tab:resolver-schema}
\end{table*}

Rows are emitted only when the winning region passes the confidence thresholds in Section~\ref{sec:lookup-confidence}. Null encodings: \texttt{top\_region} and \texttt{secondary\_region} use the empty string when no candidate region is available; \texttt{n\_content\_votes} and \texttt{content\_share} are zero for all rows in the URL-only release.

\section{Search-Time Resolution Sources}
\label{app:search-resolution-sources}

At search time, the resolver applies the priority chain in Table~\ref{tab:region-resolution-priority}. The first source that emits a valid region label determines the document's region, resolution source, and confidence value. The sources are interpreted as follows.

\begin{enumerate}
    \item \textbf{Query parameters.}
    Explicit region parameters such as \texttt{gl=US}, \texttt{country=ca}, \texttt{region=GB}, or \texttt{locale=ja-JP} receive the highest priority. These cases are rare, but they are strong signals because the site operator is explicitly encoding a locale choice.
    
    \item \textbf{Lookup-backed keys.}
    The next three sources consult the frozen lookup table. \texttt{host\_path} captures multinational websites that organize regional sites under the first path segment, such as \texttt{pwc.com/uk}. \texttt{host} captures full hostnames such as \texttt{www.lemonde.fr}. \texttt{domain} collapses subdomains to a registered domain, allowing \texttt{news.bbc.co.uk} and \texttt{sport.bbc.co.uk} to inherit evidence from \texttt{bbc.co.uk} when appropriate.
    
    \item \textbf{URL hints.}
    If the lookup table does not contain a matching key, the resolver searches for curated country tokens in structurally informative URL positions. Examples include \texttt{uk.airbnb.com}, \texttt{airbnb.com/jp/rooms}, and \texttt{redbubble.com/en-gb/}. Ambiguous tokens such as \texttt{co}, \texttt{it}, and \texttt{me} are excluded unless the URL structure makes the regional interpretation unambiguous.
    
    \item \textbf{ccTLD fallback.}
    Finally, the resolver falls back to the URL's ccTLD when it is not on a branded-ccTLD denylist. The denylist includes ccTLDs widely used outside their nominal jurisdictions, such as \texttt{.tv}, \texttt{.ly}, \texttt{.io}, \texttt{.me}, \texttt{.ai}, \texttt{.co}, \texttt{.cm}, and \texttt{.gg}. Non-branded ccTLDs such as \texttt{.de}, \texttt{.fr}, \texttt{.jp}, \texttt{.ru}, and \texttt{.br} are retained as weak fallback evidence.
    
    \item \textbf{No signal.}
    When all sources fail, the resolver emits \texttt{XX} with source \texttt{none} and confidence \texttt{none}. These records remain in the intermediate annotated corpus; filtering is left to downstream users.
\end{enumerate}

\section{Topic-to-CT Softmax and JSD}
\label{app:methods}

\paragraph{Temperature-scaled softmax for topic-to-CT projection
(Sec.~\ref{sec:corpus-cultural-topic-annotation}).}
After topic induction, each merged topic is mapped to the CT label
space. Each CT category is embedded once using BGE-M3 from its
natural-language description. For each merged topic $t$ and CT
category $c$, we compute cosine similarity between the topic centroid
$e_t$ and the CT category embedding $e_c$, and convert similarities
into a probability distribution over CT categories using a
temperature-scaled softmax:
\[
p(c \mid t) =
\frac{\exp(\cos(e_t, e_c)/\tau)}
{\sum_{c'}\exp(\cos(e_t, e_{c'})/\tau)},
\quad \tau = 0.05.
\]
The low temperature makes the mapping sharp enough for the
highest-scoring category to be interpretable, while still preserving
non-zero mass on plausible runner-up categories.

\paragraph{Jensen--Shannon distance over LTDs
(Sec.~\ref{sec:cross:locale_divergence}).}
For two locale-level Locale Topic Distributions $P, Q$ over the 14 CT
leaves, we report the Jensen--Shannon distance computed with $\log_2$:
\begin{equation*}
\begin{aligned}
\mathrm{JSD}(P, Q)
&= \sqrt{\tfrac{1}{2}\,\mathrm{KL}(P \,\|\, M)
   + \tfrac{1}{2}\,\mathrm{KL}(Q \,\|\, M)}, \\
M &= \tfrac{1}{2}(P + Q).
\end{aligned}
\end{equation*}
with $\mathrm{KL}$ in $\log_2$, so $\mathrm{JSD}(P,Q) \in [0,1]$.
Because per-locale leaf mass is produced by the temperature-scaled
softmax above, every LTD has strictly positive entries on every leaf
and no smoothing is applied; the $\mathrm{KL}$ terms are finite
without an additive prior. Implementation:
{\small\texttt{scipy.spatial.distance.jensenshannon(p,~q,~base=2)}}.

\section{Benchmark Audit Protocol}
\label{app:benchmark-audit}

This appendix describes the construction and annotation protocol for the
benchmark audit summarized in Section~\ref{sec:benchmark-audit}. The
audit is designed to support corpus--benchmark comparison by coding each
dataset along cultural coverage, scope, provenance, and usability
dimensions.

\subsection{Dataset Sources}

We compile candidate datasets from three sources:
\begin{enumerate}
    \item Datasets included in the cultural-NLP taxonomy and survey of
    \citet{liu2025culturally}.
    \item Datasets included in MUREL-style cultural-resource curation
    \citep{namazifard2025cultureneurons}.
    \item Newly identified datasets from systematic searches of the ACL
    Anthology, arXiv, HuggingFace, GitHub, and dataset repositories.
\end{enumerate}

The systematic searches combine cultural terms with resource terms.
Cultural terms include \emph{culture}, \emph{cultural},
\emph{cross-cultural}, \emph{multilingual}, \emph{norms},
\emph{values}, \emph{bias}, \emph{dialect}, \emph{stereotype},
\emph{social norms}, and \emph{cultural knowledge}. Resource terms
include \emph{dataset}, \emph{benchmark}, \emph{corpus},
\emph{evaluation}, and \emph{test set}. Candidate records are
deduplicated across source lists and benchmark families.

\subsection{Inclusion and Exclusion Criteria}

A resource is included if it satisfies all of the following criteria:
\begin{enumerate}
    \item It provides a downloadable, inspectable, or otherwise
    accessible dataset.
    \item It targets at least one cultural phenomenon covered by the CT,
    such as values, norms, dialects, communicative practices,
    culturally grounded knowledge, or culturally grounded bias.
    \item It is documented in a paper, preprint, dataset card, or
    repository with sufficient metadata for annotation.
    \item It is usable as an evaluation dataset.
\end{enumerate}

We exclude resources that do not provide dataset access, papers that
discuss cultural evaluation without releasing or identifying a dataset,
duplicate entries already represented by another variant, and resources
whose cultural target cannot be mapped to any CT leaf.

When a benchmark family contains multiple variants, we count a variant
separately only if it introduces a distinct language set, region set,
task format, annotation schema, or culturally adapted item set. For
example, a translated or culturally localised derivative is counted as
a separate dataset but linked to its parent through the lineage field.

\subsection{Annotation Dimensions}

Each included dataset is annotated along four groups of dimensions.

\paragraph{Cultural coverage.}
These fields describe what cultural phenomena the dataset tests:
\begin{itemize}
    \item CT branch;
    \item CT leaf or leaves;
\end{itemize}

\paragraph{Scope.}
These fields describe the populations, languages, and task format
covered by the dataset:
\begin{itemize}
    \item dataset name;
    \item geographical diversity;
    \item linguistic typology;
    \item languages covered;
    \item regions covered;
    \item dataset format;
    \item example item, when available.
\end{itemize}

\paragraph{Provenance.}
These fields describe where the dataset comes from and how it was
constructed:
\begin{itemize}
    \item paper link;
    \item publication venue;
    \item publication date;
    \item generation method: human, hybrid human--LLM, LLM-only,
    web-scraped or automatically extracted, or unclear;
    \item source list;
    \item dataset lineage.
\end{itemize}

\paragraph{Usability.}
These fields describe whether and how the dataset can be used for
cultural evaluation:
\begin{itemize}
    \item data link;
    \item accessibility status;
\end{itemize}

\subsection{CT Leaf Annotation}

CT branch and leaf assignments are coded from each dataset's paper,
dataset card, repository, and example items. A dataset may receive
multiple CT leaves if it evaluates more than one cultural phenomenon.
For example, a benchmark containing culturally situated dialogue about
politeness norms may be coded as both \emph{Norms \& Morals} and
\emph{Communicative Goals}; a multilingual stereotype benchmark may be
coded as \emph{Values--Bias}; and a dialect-identification benchmark is
coded as \emph{Dialects}.

Assignments are made at the dataset level rather than the individual
item level. This choice allows broad survey coverage across the
benchmark landscape, but it means that fine-grained within-dataset
variation is not fully captured. Multi-category datasets are counted
once per applicable CT leaf in the coverage matrix.

\subsection{Lineage Coding}

The lineage field records whether a dataset is original or derived from
one or more parent datasets. We mark a dataset as derived if the paper or
repository explicitly states that it is a translation, adaptation,
localisation, extension, reannotation, or culturally modified version of
an existing benchmark. Lineage is recorded using a parent-to-child
notation, for example:
\begin{quote}
\texttt{BBQ > MBBQ > KoBBQ}
\end{quote}

A dataset is coded as original if we find no known parent benchmark or
source dataset beyond generic data sources such as surveys, web pages,
or human-authored prompts. Lineage coding is conservative: if derivation
is plausible but not documented, the dataset is not marked as derived.

\subsection{Accessibility and Adaptation Coding}

Accessibility is coded according to whether the dataset can be obtained
and used for cultural evaluation:
\begin{itemize}
    \item \textbf{Public}: the dataset is available through a repository,
    dataset hub, project page, or supplementary material.
    \item \textbf{Restricted}: the dataset requires an application,
    institutional access, or approval.
    \item \textbf{Unavailable}: the dataset link is broken, missing, or
    no longer accessible.
\end{itemize}

Adaptation feasibility records whether the dataset is directly usable as
cultural assertions or evaluation items, or whether transformation is
required. Common transformations include converting QA formats,
extracting text from multimodal examples, converting classification
labels into assertions, or filtering non-cultural items from broader
multilingual datasets.

\subsection{Quality Control}

Factual metadata such as languages, regions, venue, publication date,
dataset link, and format is extracted directly from source papers,
dataset cards, or repositories. CT leaf assignments are coded by two
annotators, with disagreements resolved through discussion. The final annotated table is released with
dataset names, CT labels, language and region coverage, lineage fields,
accessibility status, and annotation notes.

\section{Robustness of the CT-Based Analyses}
\label{app:robustness}

This appendix reports the robustness analyses promised in the discussion
phase for the exploratory CT-level results: a label-perturbation study of the
Section~\ref{sec:cross:gaps} coverage classifications and the
Section~\ref{sec:cross:locale_divergence} divergence ordering, projection
behaviour inside vs.\ outside BGE-M3's documented language coverage, a noise
floor conditioned on locale size, and the decomposition behind the change in
the region-level blind-spot pattern.

\subsection{Label-Perturbation Study}
\label{app:robustness-perturbation}

The validation in Section~\ref{sec:topic-ct-validation} measures topic-level
agreement with the human consensus of 22.4\% for the adjudicated labels
(18.8\% for the embedding softmax); per-pair annotations were recorded only
in aggregate, so a full leaf-confusion matrix cannot be estimated from them.
We therefore perturb at the observed rate: each of the 801{,}344 merged
topics keeps its adjudicated leaf with probability $0.224$ and is otherwise
reassigned, drawing the replacement either uniformly over the other 13 leaves
(\emph{uniform} kernel) or proportionally to the topic's own softmax mass over
the other leaves (\emph{softmax} kernel, concentrating errors on the leaves
the projection itself confuses). Perturbed assignments enter the pipeline in
two variants: \emph{swap-soft} exchanges the softmax mass between the
original and resampled leaf within the topic's soft assignment row,
preserving the scale of the released LTDs; \emph{hard} moves the topic's
entire mass to the resampled leaf (the harshest reading, baseline: one-hot at
the adjudicated leaf). For each condition we draw $B = 1000$ replicates and
recompute the region-level 266-cell coverage, the locale-level four-class
percentages ($n = 5{,}826$ locales), and the within-language JSDs
(334{,}896 same-language locale pairs over 182 languages).

Table~\ref{tab:perturbation} summarizes the four conditions. The
region-level result of Section~\ref{sec:cross:gaps} is invariant: every
replicate of every condition keeps all 266 CT-leaf $\times$ region cells
corpus-covered at $k = 50$, with the 1st-percentile smallest cell support
across replicates at $2.6 \times 10^{5}$ documents, more than three orders
of magnitude above the threshold. Under the released soft assignment
(swap-soft), the locale-level four-class percentages are equally stable and in every replicate
all five labelled pairs of Section~\ref{sec:cross:locale_divergence} stay
above the pooled noise floor, the smallest by a factor of at least 11 at
the 2.5th percentile. Fine-grained structure is weaker: the strict
five-pair ordering survives in 44\% (uniform) and 59\% (softmax kernel) of
swap-soft replicates; the fragile comparisons are the two close ones
(en--zh preserved in 73\%/91\%, es--pt in 69\%/71\%), while the coarse
structure 
is stable, and
per-language median JSDs correlate with the unperturbed ranking at
$\rho = 0.77$--$0.83$. The hard variant additionally shows that the
locale-level percentages are tied to the soft-assignment representation
itself: replacing each topic's soft row with a one-hot at its adjudicated
leaf already shifts the unperturbed percentages to 14.5/34.0/13.9/37.6
(both / corpus-only / benchmark-only / neither), and perturbation then
moves them by up to 26 points, while the region-level coverage and the
pairs-above-floor findings continue to hold in every replicate. We
therefore read the count-based coverage results and the pairs-above-noise
finding as robust; locale-level percentages and exact pair orderings are
descriptive of the released soft-LTD pipeline, consistent with the
exploratory framing of Section~\ref{sec:topic-ct-validation}.

\begin{table*}[t]
\centering\small
\setlength{\tabcolsep}{4pt}
\begin{tabular}{@{}llccccc@{}}
\toprule
\textbf{Variant} & \textbf{Kernel} & \shortstack{\textbf{266 region cells}\\\textbf{all covered}} & \shortstack{\textbf{locale classes}\\\textbf{moved $>$2\,pt (max)}} & \shortstack{\textbf{all 5 pairs}\\\textbf{above floor}} & \shortstack{\textbf{5-pair order}\\\textbf{preserved}} & \shortstack{\textbf{Spearman $\rho$, language medians}\\\textbf{mean [95\% interval]}} \\
\midrule
swap-soft & uniform & 100\% & 0\% (0.52) & 100\% & 44\% & 0.77 [0.72, 0.81] \\
swap-soft & softmax & 100\% & 0\% (0.41) & 100\% & 59\% & 0.83 [0.79, 0.87] \\
hard & uniform & 100\% & 100\% (25.79) & 100\% & 15\% & 0.46 [0.42, 0.52] \\
hard & softmax & 100\% & 100\% (24.97) & 100\% & 15\% & 0.48 [0.44, 0.54] \\
\bottomrule
\end{tabular}
\caption{Label-perturbation study ($B=1000$ replicates per condition): every merged topic keeps its adjudicated leaf with the observed human-agreement probability 0.224 and is otherwise resampled (uniform or softmax-confusable kernel), entering the pipeline as a soft swap or a hard reassignment. Columns: fraction of replicates in which all 266 region-level cells stay corpus-covered at $k=50$; fraction in which any locale-level four-class percentage moves by more than 2 points (largest observed move, in points); fraction in which all five labelled pairs stay above the pooled $p_{95}$ noise floor; fraction preserving the full five-pair JSD ordering; Spearman correlation of per-language median JSDs (182 languages) with the unperturbed baseline.}
\label{tab:perturbation}
\end{table*}

\subsection{Projection Behaviour by BGE-M3 Coverage Regime}
\label{app:robustness-bgem3}

Table~\ref{tab:bgem3-regime} splits the topic-modeled locales by whether
their language is inside BGE-M3's documented coverage, operationalized as the
CC-100/XLM-R inventory of its backbone \citep{conneau2020unsupervised}, with
macrolanguages mapped to their standard varieties (so, e.g., Egyptian Arabic
\texttt{arz} and Cantonese \texttt{yue} count as outside). Outside documented
coverage the embedding projection is measurably less informative: the median
$\max_c p(c \mid t)$ equals the uniform value $1/14 \approx 0.071$, and 67\%
of projections are within 2\% of uniform (vs.\ 46\% inside). The
\emph{lower} adjudicator change rate outside coverage (3.5\% vs.\ 7.5\%)
should therefore not be read as better projections: near-uniform softmax rows
give the adjudicator weaker candidate rankings to correct. This quantifies
the Limitations statement that CT-level results for languages outside
BGE-M3's validated coverage carry additional uncertainty.

\begin{table}[t]
\centering\small
\setlength{\tabcolsep}{4pt}
\begin{tabular}{@{}lrr@{}}
\toprule
 & \textbf{Inside} & \textbf{Outside} \\
\midrule
Languages & 94 & 203 \\
Locales & 5{,}419 & 535 \\
Merged topics & 728{,}950 & 72{,}394 \\
Sampled documents & 27{,}922{,}687 & 1{,}772{,}582 \\
Adjudicator change rate & 7.5\% & 3.5\% \\
\quad 95\% CI (cluster bootstrap) & [7.3, 7.6] & [3.1, 3.9] \\
Softmax confidence (mean $\max_c p$) & 0.126 & 0.102 \\
Softmax confidence (median) & 0.122 & 0.071 \\
Near-uniform projections & 45.9\% & 67.0\% \\
\bottomrule
\end{tabular}
\caption{Topic-to-CT behaviour inside vs.\ outside BGE-M3's documented
language coverage (CC-100/XLM-R inventory of its backbone; standard
varieties for macrolanguages). Near-uniform: $\max_c p(c \mid t) \le
1.02/14$. The lower adjudicator change rate outside coverage is not
evidence of better projections there --- the median outside-coverage
projection carries no discriminative signal, which also weakens the
candidate rankings shown to the adjudicator.}
\label{tab:bgem3-regime}
\end{table}

\subsection{Size-Conditioned Noise Floor}
\label{app:robustness-floor}

Reviewer discussion noted that the pooled split-half noise floor of
Section~\ref{sec:cross:locale_divergence} is not conditioned on locale size.
Table~\ref{tab:size-floor} reports the same 235{,}750 split-half computations
binned by locale sample size. The floor decreases monotonically with size;
the five labelled pairs of Figure~\ref{fig:within_language_jsd} all involve
locales at the 10{,}000-document cap, where the $p_{95}$ floor is
$8.7 \times 10^{-4}$ (roughly half the pooled value) so the pooled
floor understates, rather than overstates, how far those pairs sit above
topic-modeling noise.

\begin{table}[th]
\centering\small
\setlength{\tabcolsep}{5pt}
\begin{tabular}{@{}lrccc@{}}
\toprule
 & & \multicolumn{3}{c}{\textbf{Split-half JSD} ($\times 10^{-4}$)} \\
\cmidrule(lr){3-5}
\textbf{Locale size} & \textbf{Locales} & \textbf{Median} & \textbf{IQR} & \textbf{$p_{95}$} \\
\midrule
1--2k & 1{,}047 & 10.6 & [6.2, 16.3] & 28.2 \\
2--5k & 1{,}106 & 6.6 & [3.7, 10.3] & 18.2 \\
5--10k & 650 & 4.3 & [2.3, 6.7] & 11.6 \\
10k (cap) & 1{,}912 & 3.2 & [1.6, 5.0] & 8.7 \\
\bottomrule
\end{tabular}
\caption{Split-half JSD noise floor conditioned on locale sample size
(50 splits per locale, seed 42). The floor decreases with locale size; the
five labelled pairs of Figure~\ref{fig:within_language_jsd} all lie in the
10k bin, so the pooled floor used in the main text if anything understates
their separation from noise.}
\label{tab:size-floor}
\end{table}

\subsection{English vs.\ Main-Language Evidence in the Formerly Blind Regions}
\label{app:robustness-mainlang}

Table~\ref{tab:english-vs-mainlang} decomposes corpus mass in the regions the
submitted analysis showed as blind spots. Moving from the sample-350BT
analysis corpus to the full release multiplies English-locale mass by
46--56$\times$ in every such region (the direct reversal of the
$\sim$50$\times$ English downsampling) while the mass of the regions' own
non-English main languages is unchanged, because FineWeb-2 was already
included at full scale. The region-level gap therefore closes through
English-language evidence; in the Caribbean, Oceania, and Middle Africa the
regions' own main languages contribute at most 5\% of full-release mass, and
locale-level double blind spots persist among main-language locales (e.g.,
French for Saint-Martin, \texttt{fra-MF}, 562 documents, 10 of 14 leaves
uncovered on both sides).

\begin{table*}[t]
\centering\small
\setlength{\tabcolsep}{5pt}
\begin{tabular}{@{}lrrrrrr@{}}
\toprule
 & \multicolumn{3}{c}{\textbf{English-locale mass}} & \multicolumn{2}{c}{\textbf{Non-English main-language mass (full)}} & \textbf{Main-lang.\ locales} \\
\cmidrule(lr){2-4}\cmidrule(lr){5-6}
\textbf{Region} & sample & full & growth & docs & share of region & w/ double-blind cells \\
\midrule
Caribbean & 1.4M & 66.5M & $\times$48.0 & 3.6M & 5\% & 3 \\
Oceania & 13.8M & 719.0M & $\times$52.1 & 309k & $<$1\% & 0 \\
C. America & 372k & 18.8M & $\times$50.5 & 33.9M & 60\% & 0 \\
Middle Africa & 449k & 20.5M & $\times$45.8 & 1.6M & 5\% & 0 \\
C. Asia & 85k & 4.7M & $\times$55.6 & 15.7M & 76\% & 1 \\
\bottomrule
\end{tabular}
\caption{Why the submitted region-level double-blind-spot pattern changed on
the full release, for the regions the submitted figure showed as blind.
English-locale document mass grows by $\sim$50$\times$ from the sample-350BT
analysis corpus to the full release, erasing the region-level gaps, while
non-English main-language mass is unchanged (FineWeb-2 was already at full
scale). In the Caribbean, Oceania, and Middle Africa the regions' own main
languages contribute at most 5\% of full-release mass; in Central America
and Central Asia they dominate. Locale-level double blind spots ($k=50$)
persist among main-language locales (last column).}
\label{tab:english-vs-mainlang}
\end{table*}

\section{Datasheet for FineWeb-CLaR}
\label{app:datasheet}

Following \citet{gebru2021datasheets}, we release a datasheet covering the six artefacts in our resource bundle. All artefacts share the same release URL and version-control history.

\paragraph{Motivation, composition, collection.} The corpus is annotated from the full FineWeb \citep{penedo2024fineweb} and FineWeb-2 \citep{penedo2025fineweb2} release; we do not redistribute the source text and reproduce only the metadata columns required for region attribution and topic-modeling.

\paragraph{Provenance of reported quantities.} All region-layer statistics in this paper (abstract, Section~\ref{sec:introduction}, Table~\ref{tab:region-diagnostics}) are computed on the full FineWeb + FineWeb-2 release. All cultural-topic statistics (Sections~\ref{sec:corpus-cultural-topic-annotation} and \ref{sec:experiments}) come from the released consolidated topic run over this corpus. 
Table~\ref{tab:reconciliation} reconciles the corpus snapshots from our experiments; the revised paper reports full-release values only.

\begin{table}[!th]
\centering
\scriptsize
\setlength{\tabcolsep}{2pt}
\begin{tabular}{@{}lrrr@{}}
\toprule
\textbf{Quantity} & \textbf{100BT snap.} & \textbf{350BT snap.} & \textbf{Full release} \\
\midrule
Total documents        & 5{,}176{,}421{,}725 & 5{,}547{,}240{,}607 & 30{,}914{,}158{,}759 \\
Resolved (non-XX)      & 2{,}960{,}753{,}780 & 3{,}030{,}341{,}027 & 7{,}917{,}968{,}305 \\
Resolution rate        & 57.20\%             & 54.63\%             & 25.61\% \\
Lookup-table coverage  & 1{,}636{,}856{,}362 & 1{,}681{,}071{,}710 & 4{,}627{,}776{,}301 \\
Lookup-table rate      & 31.62\%             & 30.30\%             & 14.97\% \\
\texttt{query} (reported separately) & 7{,}206{,}682 & 7{,}314{,}090 & 15{,}283{,}520 \\
\bottomrule
\end{tabular}
\caption{Reconciliation and provenance of corpus snapshots. The submitted manuscript inadvertently mixed statistics from the first two snapshots; every number in the revised paper comes from the full-release column. Each column combines the stated FineWeb subset with the full FineWeb-2.}
\label{tab:reconciliation}
\end{table}

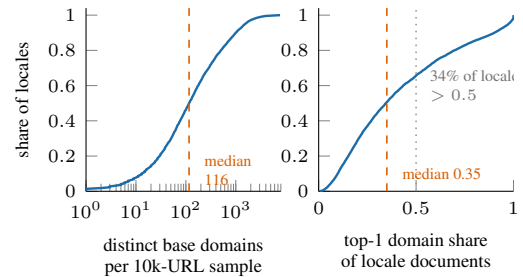
\begin{figure}[t]
    \centering
\definecolor{clsNeither}{HTML}{F0F0F0}
\definecolor{clsCorpus}{HTML}{56B4E9}
\definecolor{clsSurvey}{HTML}{E69F00}
\definecolor{clsBoth}{HTML}{004488}
\definecolor{OIblue}{HTML}{0072B2}
\definecolor{medblue}{HTML}{1B6CA8}
\definecolor{lightblue}{HTML}{A6CBE3}
\definecolor{oiorange}{HTML}{D95F02}
\definecolor{histblue}{HTML}{4A87E0}
\definecolor{darkblue}{HTML}{123E85}
\begin{tikzpicture}
\begin{groupplot}[
  group style={group size=2 by 1, horizontal sep=0.5cm, ylabels at=edge left},
  width=0.54\linewidth, height=0.52\linewidth,
  ymin=0, ymax=1.04, ytick={0,0.2,0.4,0.6,0.8,1.0},
  axis x line*=bottom, axis y line*=left,
  ylabel={share of locales},
  label style={font=\scriptsize},
  tick label style={font=\scriptsize},
  xlabel style={align=center, font=\scriptsize},
]
\nextgroupplot[xmode=log, xmin=1, xmax=8000,
  xlabel={distinct base domains\\per 10k-URL sample}]
\addplot[medblue, line width=0.9pt] coordinates { (1,0.0002) (1,0.0047) (1,0.0092) (1,0.0138) (2,0.0183) (3,0.0228) (3,0.0275) (4,0.0321) (5,0.0366) (5,0.0411) (6,0.0457) (6,0.0504) (7,0.0549) (7,0.0595) (8,0.064) (9,0.0685) (9,0.0731) (10,0.0778) (11,0.0823) (11,0.0868) (12,0.0914) (13,0.0959) (13,0.1006) (14,0.1051) (15,0.1097) (15,0.1142) (16,0.1187) (16,0.1233) (17,0.128) (18,0.1325) (19,0.1371) (20,0.1416) (20,0.1461) (21,0.1508) (22,0.1554) (23,0.1599) (24,0.1644) (24,0.169) (25,0.1735) (26,0.1782) (27,0.1827) (27,0.1873) (28,0.1918) (29,0.1963) (30,0.201) (31,0.2056) (31,0.2101) (32,0.2146) (33,0.2192) (34,0.2237) (35,0.2284) (36,0.233) (37,0.2375) (38,0.242) (38,0.2466) (39,0.2513) (40,0.2558) (41,0.2603) (42,0.2649) (43,0.2694) (44,0.2739) (45,0.2786) (47,0.2832) (48,0.2877) (49,0.2922) (50,0.2968) (51,0.3015) (52,0.306) (53,0.3105) (54,0.3151) (55,0.3196) (56,0.3242) (57,0.3289) (58,0.3334) (59,0.3379) (61,0.3425) (62,0.347) (63,0.3517) (64,0.3562) (65,0.3608) (66,0.3653) (67,0.3698) (68,0.3744) (70,0.3791) (72,0.3836) (73,0.3881) (74,0.3927) (76,0.3972) (77,0.4019) (79,0.4064) (80,0.411) (82,0.4155) (83,0.4201) (85,0.4246) (87,0.4293) (88,0.4338) (90,0.4384) (92,0.4429) (93,0.4474) (95,0.4521) (96,0.4567) (99,0.4612) (101,0.4657) (103,0.4703) (104,0.4748) (107,0.4795) (109,0.484) (111,0.4886) (113,0.4931) (115,0.4976) (117,0.5024) (120,0.5069) (121,0.5114) (124,0.516) (128,0.5205) (130,0.5252) (132,0.5297) (134,0.5343) (137,0.5388) (140,0.5433) (142,0.5479) (145,0.5526) (148,0.5571) (150,0.5616) (152,0.5662) (155,0.5707) (158,0.5754) (161,0.5799) (165,0.5845) (168,0.589) (170,0.5936) (174,0.5981) (179,0.6028) (182,0.6073) (186,0.6119) (189,0.6164) (194,0.6209) (198,0.6256) (202,0.6302) (207,0.6347) (210,0.6392) (214,0.6438) (218,0.6483) (222,0.653) (227,0.6575) (232,0.6621) (239,0.6666) (242,0.6711) (249,0.6758) (257,0.6804) (263,0.6849) (271,0.6895) (278,0.694) (285,0.6985) (293,0.7032) (299,0.7078) (306,0.7123) (309,0.7168) (320,0.7214) (327,0.7261) (333,0.7306) (344,0.7351) (353,0.7397) (362,0.7442) (372,0.7487) (379,0.7534) (388,0.758) (397,0.7625) (407,0.767) (417,0.7716) (430,0.7763) (443,0.7808) (462,0.7854) (475,0.7899) (488,0.7944) (499,0.799) (517,0.8037) (533,0.8082) (547,0.8127) (560,0.8173) (582,0.8218) (599,0.8265) (622,0.831) (641,0.8356) (663,0.8401) (680,0.8446) (695,0.8492) (723,0.8539) (748,0.8584) (773,0.8629) (795,0.8675) (822,0.872) (849,0.8767) (880,0.8813) (912,0.8858) (938,0.8903) (960,0.8949) (1003,0.8994) (1032,0.9041) (1069,0.9086) (1106,0.9132) (1154,0.9177) (1209,0.9222) (1259,0.9269) (1317,0.9315) (1364,0.936) (1407,0.9405) (1469,0.9451) (1536,0.9496) (1607,0.9543) (1698,0.9589) (1764,0.9634) (1879,0.9679) (2024,0.9725) (2226,0.9772) (2458,0.9817) (2749,0.9862) (3207,0.9908) (3975,0.9953) (7712,1.0) };
\draw[oiorange, dashed, line width=0.6pt]
  (axis cs:116,0) -- (axis cs:116,1.04);
\node[font=\tiny, oiorange, anchor=west, align=left]
  at (axis cs:116,0.12) [xshift=2pt] {median\\116};
\nextgroupplot[xmin=0, xmax=1, xtick={0,0.5,1},
  xlabel={top-1 domain share\\of locale documents}]
\addplot[medblue, line width=0.9pt] coordinates { (0.0061,0.0002) (0.0207,0.0047) (0.0277,0.0092) (0.0336,0.0138) (0.0385,0.0183) (0.0441,0.0228) (0.0476,0.0275) (0.0516,0.0321) (0.0555,0.0366) (0.0593,0.0411) (0.0625694,0.0457) (0.0662,0.0504) (0.0694,0.0549) (0.0724638,0.0595) (0.0764415,0.064) (0.0794,0.0685) (0.0819062,0.0731) (0.0852535,0.0778) (0.0878,0.0823) (0.0904,0.0868) (0.0933,0.0914) (0.0955577,0.0959) (0.0979689,0.1006) (0.1006,0.1051) (0.1029,0.1097) (0.104683,0.1142) (0.1067,0.1187) (0.1091,0.1233) (0.1112,0.128) (0.11375,0.1325) (0.116,0.1371) (0.1185,0.1416) (0.121372,0.1461) (0.1238,0.1508) (0.127004,0.1554) (0.1298,0.1599) (0.13245,0.1644) (0.134109,0.169) (0.136754,0.1735) (0.139659,0.1782) (0.142416,0.1827) (0.1453,0.1873) (0.147595,0.1918) (0.150129,0.1963) (0.1528,0.201) (0.155031,0.2056) (0.15711,0.2101) (0.160116,0.2146) (0.162206,0.2192) (0.164853,0.2237) (0.167577,0.2284) (0.1699,0.233) (0.172583,0.2375) (0.1742,0.242) (0.176639,0.2466) (0.1799,0.2513) (0.182009,0.2558) (0.184759,0.2603) (0.1869,0.2649) (0.190115,0.2694) (0.192423,0.2739) (0.194434,0.2786) (0.1964,0.2832) (0.198623,0.2877) (0.201191,0.2922) (0.203883,0.2968) (0.2063,0.3015) (0.209089,0.306) (0.2124,0.3105) (0.214888,0.3151) (0.217039,0.3196) (0.2206,0.3242) (0.22285,0.3289) (0.224687,0.3334) (0.227633,0.3379) (0.230098,0.3425) (0.23356,0.347) (0.236854,0.3517) (0.239766,0.3562) (0.243,0.3608) (0.246316,0.3653) (0.249813,0.3698) (0.2522,0.3744) (0.255462,0.3791) (0.258442,0.3836) (0.261438,0.3881) (0.265068,0.3927) (0.268181,0.3972) (0.2711,0.4019) (0.273643,0.4064) (0.2769,0.411) (0.279783,0.4155) (0.282357,0.4201) (0.285608,0.4246) (0.288998,0.4293) (0.29329,0.4338) (0.296733,0.4384) (0.299844,0.4429) (0.302745,0.4474) (0.306228,0.4521) (0.308873,0.4567) (0.3137,0.4612) (0.3174,0.4657) (0.319777,0.4703) (0.325192,0.4748) (0.328627,0.4795) (0.3329,0.484) (0.336889,0.4886) (0.339261,0.4931) (0.342395,0.4976) (0.346833,0.5024) (0.350419,0.5069) (0.354658,0.5114) (0.357673,0.516) (0.3615,0.5205) (0.364979,0.5252) (0.368633,0.5297) (0.374257,0.5343) (0.378225,0.5388) (0.381579,0.5433) (0.3851,0.5479) (0.3888,0.5526) (0.3918,0.5571) (0.396472,0.5616) (0.402005,0.5662) (0.407882,0.5707) (0.410455,0.5754) (0.4158,0.5799) (0.4198,0.5845) (0.4244,0.589) (0.428952,0.5936) (0.432545,0.5981) (0.436108,0.6028) (0.44197,0.6073) (0.4462,0.6119) (0.454446,0.6164) (0.460081,0.6209) (0.4657,0.6256) (0.471052,0.6302) (0.476502,0.6347) (0.4819,0.6392) (0.487833,0.6438) (0.4919,0.6483) (0.496662,0.653) (0.5004,0.6575) (0.5048,0.6621) (0.510204,0.6666) (0.5165,0.6711) (0.524034,0.6758) (0.529045,0.6804) (0.533221,0.6849) (0.539984,0.6895) (0.54641,0.694) (0.550725,0.6985) (0.555726,0.7032) (0.561497,0.7078) (0.56705,0.7123) (0.573372,0.7168) (0.5777,0.7214) (0.5825,0.7261) (0.589773,0.7306) (0.594591,0.7351) (0.6002,0.7397) (0.605263,0.7442) (0.61168,0.7487) (0.618823,0.7534) (0.626073,0.758) (0.635054,0.7625) (0.643973,0.767) (0.6528,0.7716) (0.6582,0.7763) (0.665574,0.7808) (0.673333,0.7854) (0.680851,0.7899) (0.686308,0.7944) (0.694308,0.799) (0.705194,0.8037) (0.7124,0.8082) (0.720485,0.8127) (0.728889,0.8173) (0.736721,0.8218) (0.745008,0.8265) (0.7548,0.831) (0.762281,0.8356) (0.770784,0.8401) (0.781,0.8446) (0.789474,0.8492) (0.8004,0.8539) (0.8092,0.8584) (0.8224,0.8629) (0.8324,0.8675) (0.8434,0.872) (0.853801,0.8767) (0.865036,0.8813) (0.874346,0.8858) (0.882939,0.8903) (0.893214,0.8949) (0.900937,0.8994) (0.908815,0.9041) (0.916058,0.9086) (0.923602,0.9132) (0.9291,0.9177) (0.935878,0.9222) (0.9436,0.9269) (0.950226,0.9315) (0.9567,0.936) (0.962869,0.9405) (0.9696,0.9451) (0.974,0.9496) (0.978052,0.9543) (0.983645,0.9589) (0.986469,0.9634) (0.990183,0.9679) (0.993816,0.9725) (0.996795,0.9772) (0.998319,0.9817) (1,0.9862) (1,0.9908) (1,0.9953) (1,1.0) };
\draw[oiorange, dashed, line width=0.6pt]
  (axis cs:0.35,0) -- (axis cs:0.35,1.04);
\draw[black!40, dotted, line width=0.6pt] (axis cs:0.5,0) -- (axis cs:0.5,1.04);
\node[font=\tiny, oiorange, anchor=west]
  at (axis cs:0.35,0.10) [xshift=2pt] {median 0.35};
\node[font=\tiny, black!55, anchor=west, align=left]
  at (axis cs:0.52,0.60) {34\% of locales\\$> 0.5$};
\end{groupplot}
\end{tikzpicture}
    \caption{URL base-domain diversity per topic-modeled locale (up to
    10{,}000 URLs sampled per locale from the full release). Left: CDF of
    distinct registered base domains per locale (median 116). Right: CDF of
    the top-1 domain's share of locale documents (median 0.35); 34\% of
    locales draw more than half of their documents from a single domain.}
    \label{fig:url_domain_diversity}
\end{figure}

\paragraph{PII and query-string filtering.} The lookup table is keyed on \texttt{key\_type} $\in \{$\texttt{domain}, \texttt{host}, \texttt{host\_path}$\}$; full URLs and query strings are never persisted in the lookup table. At search time, the resolver inspects query strings only to extract explicit locale parameters (\texttt{gl}, \texttt{country}, \texttt{region}, \texttt{locale}); other query parameters are discarded before annotation columns are written. No personal data, session tokens, or user identifiers are emitted in the released annotations.

\begin{table}[!th]
\centering
\scriptsize
\begin{tabular}{@{}p{0.48\linewidth}p{0.22\linewidth}p{0.22\linewidth}@{}}
\toprule
\textbf{Artifact} & \textbf{Format} & \textbf{License} \\
\midrule
Region-annotated corpus & Parquet shards & ODC-By 1.0 \\
URL-signature lookup table & CSV & CC-BY 4.0 \\
Locale Topic Distributions & Parquet & CC-BY 4.0 \\
Per-document culture annotations & Parquet & ODC-By 1.0 \\
Benchmark survey & BibTeX + CSV & CC-BY 4.0 \\
Pipeline code & Git repository & Apache-2.0 \\
\bottomrule
\end{tabular}
\caption{Released artifacts, formats, and licenses. The region-annotated corpus covers the full FineWeb + FineWeb-2 release (30.9B documents) and inherits their ODC-By 1.0 license. Locale Topic Distributions cover 5{,}954 locales, and the benchmark survey contains 277 entries.}
\label{tab:datasheet-artifacts}
\end{table}

\paragraph{Distribution, maintenance.} Artefacts are released under the licences in Table~\ref{tab:datasheet-artifacts} with a fixed version tag; future versions will be released alongside a changelog. The region-annotated corpus is distributed as \texttt{Yusser/FineWeb-CLaR-region} (locale-partitioned, id-joinable to the source corpora) and the culture-annotated locale samples as \texttt{Yusser/FineWeb-CLaR-culture} (with per-document topic and CT metadata), both on the Hugging Face Hub.

\section{Compute and Wall-Clock Disclosure}
\label{app:compute}

All region-attribution runs were single-machine, CPU-bound, with 64 worker processes against the frozen lookup table.

\begin{figure}[th]
    \centering
\definecolor{clsNeither}{HTML}{F0F0F0}
\definecolor{clsCorpus}{HTML}{56B4E9}
\definecolor{clsSurvey}{HTML}{E69F00}
\definecolor{clsBoth}{HTML}{004488}
\definecolor{OIblue}{HTML}{0072B2}
\definecolor{medblue}{HTML}{1B6CA8}
\definecolor{lightblue}{HTML}{A6CBE3}
\definecolor{oiorange}{HTML}{D95F02}
\definecolor{histblue}{HTML}{4A87E0}
\definecolor{darkblue}{HTML}{123E85}
\begin{tikzpicture}
\begin{axis}[
  width=1.0\linewidth, height=0.68\linewidth,
  ymin=0, ymax=348, xmin=-4, xmax=204,
  xtick={0,50,100,150,200},
  axis x line*=bottom, axis y line*=left,
  xlabel={merged topics per locale (K=200 budget)},
  ylabel={locales},
  label style={font=\scriptsize},
  tick label style={font=\scriptsize},
  legend style={at={(0.04,0.96)}, anchor=north west, draw=none, fill=none, font=\scriptsize},
  legend cell align=left,
]
\addplot[ybar interval, fill=histblue, draw=white, line width=0.3pt, forget plot]
  coordinates { (0,14) (5,21) (10,32) (15,14) (20,10) (25,9) (30,4) (35,9) (40,14) (45,15) (50,32) (55,24) (60,38) (65,49) (70,65) (75,101) (80,126) (85,151) (90,156) (95,211) (100,236) (105,257) (110,239) (115,218) (120,205) (125,190) (130,233) (135,254) (140,296) (145,308) (150,291) (155,325) (160,282) (165,286) (170,264) (175,286) (180,288) (185,238) (190,132) (195,31) (200,0) };
\addplot[oiorange, dashed, line width=0.9pt] coordinates {(134.6,0) (134.6,348)};
\addlegendentry{mean 135}
\addplot[darkblue, dotted, line width=1.0pt] coordinates {(140,0) (140,348)};
\addlegendentry{median 140}
\end{axis}
\end{tikzpicture}
    \caption{Merged topics per locale across the 5{,}954 topic-modeled
    locales (801{,}344 in total). Per-locale merging of the $K=200$ FASTopic
    budget at cosine $0.85$ retains an effective average of
    $\approx$135 topics per locale.}
    \label{fig:merged-topic-counts}
\end{figure}
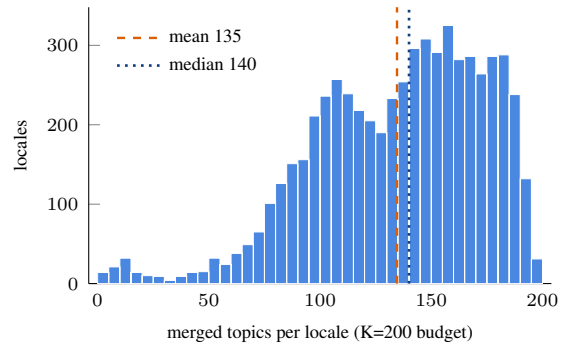

\paragraph{Topic modeling (BGE-M3 + FASTopic).} Per-locale topic
induction over the 5{,}954 topic-modeled locales used multilingual
BGE-M3 \citep{chen2024m3} for document encoding and FASTopic
\citep{wu2024fastopic} for topic induction with $K=200$ initial
topics per locale. Document embedding ran in two passes: an initial
pass on 6$\times$ NVIDIA A100 80\,GB GPUs ($\approx$65\,GPU-h,
8{,}192-token inputs), whose per-locale embeddings were reused, and a
consolidation pass on single NVIDIA H100 GPUs ($\approx$1.0\,GPU-h,
1{,}024-token cap) covering the locales fetched or re-sampled for the
final locale set. Topic induction for the final run took
$\approx$0.7\,GPU-h with per-locale crash-resume checkpointing, and
CT projection with LLM adjudication dominated the cost at
$\approx$273\,GPU-h (split across two parallel single-GPU jobs), plus
$\approx$0.8\,CPU-h for the aggregate analysis.
Figure~\ref{fig:merged-topic-counts} shows the distribution of merged-topic
counts per locale under the $K=200$ budget (mean 135, median 140,
5th--95th percentile 70--187; e.g.\ ar-EG 106, en-US 84, nya-MW 154).

\paragraph{LLM adjudication.} Low-confidence topic-to-CT projections
($\max_c p(c \mid t) < 0.5$) were adjudicated by
Qwen2.5-32B-Instruct (AWQ-quantised, served via vLLM~0.19.0 with
\texttt{enforce\_eager=True}) at temperature $T = 0$. Because the
temperature-scaled softmax at $\tau = 0.05$ over the CT label space
produces a flat maximum in practice (observed $\max_c p(c \mid t)$:
mean $0.124$, 99th percentile $0.315$), 801{,}068 of the 801{,}344
merged topics across the 5{,}954 locales (99.97\%) fell below the
$0.5$ cutoff and were adjudicated; the adjudicator changed the
primary CT-leaf assignment for 7.1\% of adjudicated topics
(56{,}931).


\begin{table*}[h]
\centering
\tiny
\setlength{\tabcolsep}{4pt}
\begin{tabular}{p{2.8cm}p{12.5cm}}
\toprule
\textbf{Element} & \textbf{Datasets} \\
\midrule
\multicolumn{2}{l}{\textbf{\textit{Ideational}}} \\[2pt]
  Concepts (21) & \citet{hu-etal-2024-bridging, cao2023_231017353, nayak-etal-2024-benchmarking, romero2024_240605967, winata-etal-2025-worldcuisines, myung2024_240609948, nguyen2023candle, guo2025_250405154, majewska2022_220113405, culturegen2024}; \citet{li2024_240805102, yin-etal-2021-broaden, koto-etal-2024-indoculture, liu-etal-2021-visually, hu-etal-2023-multi-3, wang2021_210301910, ayash2025_250317485, lee2023viscounth, thapliyal-etal-2022-crossmodal, li-etal-2023-pace}; \citet{silknow2022} \\[3pt]
  Knowledge (32) & \citet{romero2024_240605967, koto2024_240212840, myung2024_240609948, bnmmlu2026, nguyen2023candle, li-etal-2024-cmmlu, wibowo-etal-2024-copal, chiu2024_241002677, culturegen2024, shi-etal-2024-culturebank}; \citet{keleg-magdy-2023-dlama, hardalov-etal-2020-exams, semenov-sennrich-2025-measuring, yin-etal-2022-geomlama, singh-etal-2025-global, koto-etal-2024-indoculture, koto-etal-2023-large, kmmlu2024, fitzgerald2022_220408582, sakai-etal-2024-mcsqa}; \citet{lin-etal-2021-common, kassner-etal-2021-multilingual, xia-monti-2021-multilingual, pramodya-etal-2025-translating, jiang-etal-2020-x, ponti-etal-2020-xcopa, yuan-etal-2024-measuring, maps2024, aya2024_240206619, xstest2024_230801263}; \citet{mlqa2020_191007475, tydi2020_200305002} \\[3pt]
  Values - general (15) & \citet{scaria-etal-2024-instructabsa, li2024_240515145, prism2024, aakanksha2024_240618682, pistilli2024civics, karinshak2024_241106032, zahraei2025_251013154, xu-etal-2024-exploring-multilingual, cahyawijaya-etal-2025-high, li2024_240210946}; \citet{globalopinionqa2024, zhao-etal-2024-worldvaluesbench, valueprism2024, seadialogues2025_250807069, global2004} \\[3pt]
  Values - bias (28) & \citet{saralegi-zulaika-2025-basqbbq, joshi2025_250801710, ravikiran-annamalai-2021-dosa, palta-rudinger-2023-fork, nozza-etal-2021-honest, multilingualbia2024, multitaskchines2025, bhutani-etal-2024-seegull, mitchell-etal-2025-shades, nadeem-etal-2021-stereoset}; \citet{ozturk2023_230707331, troles-schmid-2021-extending, stanovsky-etal-2019-evaluating, kappl2025_250219104, xu-etal-2024-exploring-multilingual, lauscher-etal-2020-araweat, parrish-etal-2022-bbq, nangia-etal-2020-crows, strazda2025_250716442, neveol-etal-2022-french}; \citet{sahoo-etal-2024-indibias, jin2023_230716778, mukherjee-etal-2023-global, zhao2018_180406876, lauscher2019_190411783, tomar2025_250807090, winogender2018_180409301, xstest2024_230801263} \\[3pt]
  Values - hate (13) & \citet{joshi2025_250801710, ravikiran-annamalai-2021-dosa, mandl2021_210805927, bassignana-etal-2018-hurtlex, trager-etal-2025-mftcxplain, luu2021_210311528, lee2023_230816705, vargas-etal-2022-hatebr, vargas2026_260103481, mathew2020_201210289}; \citet{bui-etal-2025-multi3hate, dementieva-etal-2025-multilingual, wikipediatoxicitysubtypes2017_161008914} \\[3pt]
  Values - other perceptions (12) & \citet{mohamed-etal-2022-artelingo, mohamed-etal-2024-culture, belay2025_250310688, frenda-etal-2023-epic, havaldar-etal-2024-building, deas2024_240712196, pei-etal-2023-semeval, casola-etal-2024-multipico, yue-etal-2024-sarcnet, barriere-etal-2025-standup4ai}; \citet{hopeedi2021, kanhope2021_210804616} \\[3pt]
  Norms and Morals (36) & \citet{chen2023cmfd, guo2025_250405154, wang-etal-2024-cdeval, joshi2025_250801710, li2024_240515145, trager-etal-2025-mftcxplain, zhan-etal-2024-renovi, lee2023socialdial, xu-etal-2024-exploring-multilingual, cahyawijaya-etal-2025-high}; \citet{vargas2026_260103481, huang-yang-2023-culturally, pyatkin2022_221210409, yu-etal-2024-cmoraleval, socnormnli2023, ohashi2024_241009564, dwivedi-etal-2023-eticor, dwivedi-etal-2025-eticor, hopp2021emfd, ziems-etal-2022-moral}; \citet{sahu-etal-2025-minds, hoover2020mftc, emelin-etal-2021-moral, rao2024_240412464, ziems-etal-2023-normbank, li-etal-2023-normdial, fung-etal-2023-normsage, kim-etal-2022-prosocialdialog, saffari2024_240609123, yuan-etal-2024-measuring}; \citet{forbes-etal-2020-social, hobson-etal-2024-story, valueprism2024, moralconvita2021, seadialogues2025_250807069} \\[3pt]
  Artifacts (12) & \citet{mohamed-etal-2022-artelingo, mohamed-etal-2024-culture, koto-etal-2024-indoculture, hossain-etal-2022-memosen, tikhonov-etal-2021-storydb, lee2023viscounth, hobson-etal-2024-story, epure-etal-2020-modeling, bwb_discourse2023, seadialogues2025_250807069}; \citet{euronews2016, midas2020} \\[3pt]
\midrule
\multicolumn{2}{l}{\textbf{\textit{Linguistic Elements}}} \\[2pt]
  Dialects (15) & \citet{winata-etal-2025-worldcuisines, wibowo-etal-2024-copal, frenda-etal-2023-epic, singh-etal-2025-global, casola-etal-2024-multipico, dementieva-etal-2025-multilingual, ziems-etal-2023-multi, dialect2standard2023, dialectbench2024, divsumm2022}; \citet{twt20232023, sds2002022, stt4sg3502023, vernacular2022_220403031, aya2024_240206619} \\[3pt]
  Styles/Reg./Genres (16) & \citet{winata-etal-2025-worldcuisines, guo2025_250405154, sweed-shahaf-2021-catchphrase, nadejde-etal-2022-cocoa, krishna-etal-2022-shot, khoshtab-etal-2025-comparative, fitzgerald2022_220408582, lee2023socialdial, srinivasan-choi-2022-tydip, briakou-etal-2021-ola}; \citet{li-etal-2023-pace, laippala-etal-2022-towards, sun-xu-2022-tracing, kuzman2023, holistic2023, mtst2023} \\[3pt]
\midrule
\multicolumn{2}{l}{\textbf{\textit{Social Elements}}} \\[2pt]
  Relationship (7) & \citet{wang-etal-2024-cdeval, chiu2024_241002677, zhan-etal-2024-renovi, lee2023socialdial, li-etal-2023-normdial, culturalcodes2023, global2004} \\[3pt]
  Context (17) & \citet{wang-etal-2024-cdeval, ravikiran-annamalai-2021-dosa, wang2020_201108772, fitzgerald2022_220408582, currey-etal-2022-mt, wang2021_210301910, pramodya-etal-2025-translating, goel-etal-2023-presto, zhan-etal-2024-renovi, lee2023socialdial}; \citet{socnormnli2023, emelin-etal-2021-moral, ziems-etal-2023-normbank, kim-etal-2022-prosocialdialog, forbes-etal-2020-social, laippala-etal-2022-towards, kuzman2023} \\[3pt]
  Communicative Goals (11) & \citet{wang-etal-2024-cdeval, goel-etal-2023-presto, zhan-etal-2024-renovi, lee2023socialdial, emelin-etal-2021-moral, ziems-etal-2023-normbank, li-etal-2023-normdial, laippala-etal-2022-towards, kuzman2023, jin-etal-2024-persuading}; \citet{global2004} \\[3pt]
  Demographics (12) & \citet{ava2024, frenda-etal-2023-epic, trager-etal-2025-mftcxplain, pei-etal-2023-semeval, casola-etal-2024-multipico, prism2024, lee2023_230816705, ziems-etal-2023-normbank, saffari2024_240609123, zhao-etal-2024-worldvaluesbench}; \citet{culturalcodes2023, stt4sg3502023} \\[3pt]
\bottomrule
\end{tabular}
\caption{Datasets organized by taxonomy element, following \citet{liu2025culturally}. Numbers in parentheses indicate dataset count per category. Multi-category datasets appear under each applicable element.}
\label{tab:datasets}
\end{table*}

\end{document}